%% file: main.tex
\documentclass[10pt,twocolumn,letterpaper]{article}

\usepackage[pagenumbers]{cvpr}%

\input{preamble}
\definecolor{cvprblue}{rgb}{0.21,0.49,0.74}
\usepackage[pagebackref,breaklinks,colorlinks,allcolors=cvprblue]{hyperref}

\def\confName{CVPR}
\def\confYear{2027}

\newcommand{\dataset}{HOT3D-Grasp}
\newcommand{\method}{GRAZE}
\newcommand{\texthoi}{Text2HOI}
\newcommand{\geogaze}{GeoGaze}
\newcommand{\yesmark}{\textcolor{green!55!black}{\ensuremath{\checkmark}}}
\newcommand{\nomark}{\textcolor{red!70!black}{\ensuremath{\times}}}
\newif\ifwithsupp \withsupptrue%
\newcommand{\suppref}[2]{\ifwithsupp\ref{#1}\else#2\fi}

\title{GRAZE: Text-Free Grasp Synthesis for Unseen Objects from Human Gaze}

\author{Jeongho Lee$^{1}$ \quad Boeun Kim$^{1}$ \quad Hyung Jin Chang$^{2}$ \quad Jungchan Cho$^{3}$ \quad Sang-Il Choi$^{1}$\\[4pt]
$^{1}$Dankook University \quad $^{2}$University of Birmingham \quad $^{3}$Gachon University
}

\begin{document}
\maketitle
\input{sec/0_abstract}
\input{sec/1_intro}
\input{sec/2_related}
\input{sec/3_method}

\input{sec/4_experiments}

\input{sec/5_conclusion}
\input{sec/6_llm_statement}
{
    \small
    \bibliographystyle{ieeenat_fullname}
    \bibliography{main}
}

\ifwithsupp
\clearpage
\input{sec/X_suppl}
\fi

\end{document}

%% file: preamble.tex
\usepackage{cuted}%
\usepackage{array}%
\ifdefined\pdfsuppressptexinfo \pdfsuppressptexinfo=-1 \fi%
\ifdefined\pdfinfoomitdate \pdfinfoomitdate=1 \fi

%% file: sec/0_abstract.tex
\begin{strip}
\centering
\includegraphics[width=\textwidth]{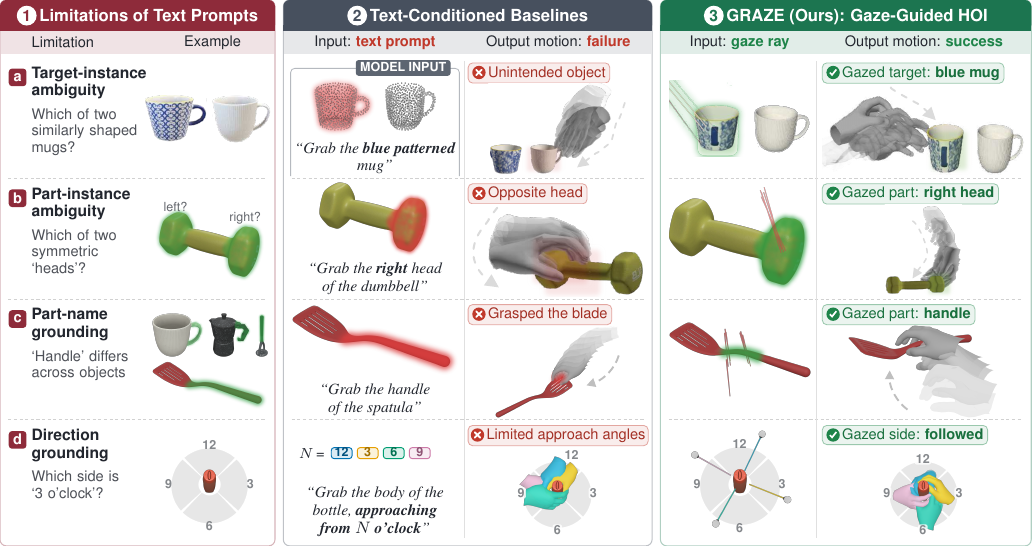}
\captionof{figure}{\textbf{Where text prompts fall short.} Text baselines receive object and part names. \method{} receives gaze instead. (a)~Asked for ``the blue patterned mug'', a text baseline grasps the similarly shaped white mug. (b)~With ``left'' or ``right'', a text baseline grasps the same end of the dumbbell. (c)~Asked for the spatula's handle, a text baseline grasps the blade, while \method{} grasps the gazed handle. (d)~A text baseline rarely turns its approach when the clock word changes, but \method{} turns with the rotated gaze. Output cells show motions generated by each model on test clips of unseen objects (faded hands: earlier frames).}
\label{fig:task_overview}
\vspace{2mm}
\end{strip}

\begin{abstract}
Text-conditioned hand--object interaction (HOI) generation provides intuitive semantic control, but language is an indirect way to specify where an interaction should occur on an object. Naming the part to grasp requires object-specific words, and naming the approach side requires a reference frame that a prompt lacks. We propose \textbf{\method{}} (GRAsp from gaZE), which generates hand--object grasp motion from recorded gaze instead of object and part names. Its \textbf{Geometry-Aware Gaze Encoding (\geogaze{})} maps gaze rays to cues on object geometry, and cross-attention conditions a diffusion Transformer on these cues.
We also introduce \textbf{\dataset{}}, a grasp-centered benchmark with object-disjoint folds built on HOT3D recordings. On unseen objects, \method{} grasps the target part more often than the strongest text-conditioned baseline (49.3\% vs.\ 43.6\%), achieves the highest grasp success (66.3\% vs.\ 46.6\%), and transfers without fine-tuning to GazeHOI (45.3\% vs.\ 33.8\%). Gaze conveys the approach side: when gaze is rotated around the object, the approach turns with it in 59.1\% of cases, against 20.5\% for a text model retrained with clock words. On the same backbone, a text prompt names parts better, but gaze adds the approach side. Given both, \method{} keeps the prompt's part accuracy and follows gaze.
\end{abstract}

%% file: sec/1_intro.tex
\section{Introduction}
\label{sec:intro}

Generating realistic hand--object interaction (HOI) motion is essential for animating hands in virtual and mixed reality. Recent generative models synthesize HOI motion from a text prompt~\citep{text2hoi,diffh2o,latenthoi,hoigpt,jointhoi,hoflow,hoi-diff,thor}. Earlier methods needed the wrist or object trajectory as input~\citep{manipnet}, whereas a prompt only has to describe what the hand should do, such as ``grab the handle of the mug.''

Yet language is not always the most direct control signal for grasping. Text conveys \emph{semantic intention}, such as the object category and the action, but grasping also requires \emph{spatial intention}: which object to approach and where to contact it. Text-conditioned methods such as \texthoi{} infer contact from the prompt and the object geometry, so the user has no direct way to specify the grasp location~\citep{text2hoi}. Words fall short in three cases. First, these models observe objects only as geometry, so an appearance word cannot separate similarly shaped objects (\cref{fig:task_overview}a). Second, a part name such as a dumbbell's ``head'' may refer to two symmetric instances, and ``left'' or ``right'' cannot pick one because these words depend on an unstated viewpoint: the same head is on the left from one side of the table and on the right from the other (\cref{fig:task_overview}b). Third, the same part name (\eg, handle) denotes differently shaped regions on different objects, so the prompt does not identify the intended grasp location (\cref{fig:task_overview}c).

These cases concern where to grasp. Text also leaves the approach side unspecified. This side matters in real scenes: if the handle of a mug faces away from the person, the hand must reach around the mug or first turn it, so a plausible grasp motion depends on how the object is rotated relative to the person. The object can lie in any pose relative to the person, but text-conditioned methods never observe where the person stands. Adding a direction word does not help: a user's ``3 o'clock'' has no reference frame in the scene, and even a text model retrained with object-frame clock words turns its approach to match a changed word no more often than a random direction (25\%, Figs.~\ref{fig:task_overview}d and~\ref{fig:control}b).

These limitations motivate an input that needs no object or part names and carries spatial information that words express poorly, such as where the person is relative to the object. Human gaze provides such an input: people typically look toward an intended object or region before and during interaction~\citep{gaze_3,gaze_1,gaze_2,gaze_4,gaze_6}. Gaze does not define a complete hand trajectory, but it offers two spatial cues for grasp generation: where the ray meets the object marks the intended part, and where the ray starts, at the eye, tells which side of the object faces the person. Gaze is also cheap to obtain: the eye tracker of an egocentric headset records it without annotation.

In this paper, we propose \method{} (GRAsp from gaZE), a gaze-guided framework for generating
grasp-based HOI motion toward the object region the person looks at. Given the object's geometry, its initial pose, and the first 30\% of a recorded gaze sequence, \method{} synthesizes one grasp-and-lift motion offline. Our encoder \geogaze{} maps
each gaze ray onto fixed Basis Point Set (BPS) points~\citep{bps}, using ray closeness for local proximity
and displacement alignment for global direction. Cross-attention passes these cues to a diffusion Transformer. MANO fitting~\citep{mano} and a contact refinement then turn the predicted hand vertices into grasps.

Our contributions are threefold. First, we formulate \textbf{text-free, gaze-guided grasp-based HOI generation}, in which recorded gaze replaces object and part names as the condition. On unseen objects, \method{} grasps the target part more often than the strongest text baseline, especially on handles, which the text baselines rarely grasp even when the handle is named (Figs.~\ref{fig:task_overview}c and~\ref{fig:convey}a). Gaze also conveys the approach side, which text leaves open: when the gaze is rotated around the object, the approach of \method{} turns with it in 59.1\% of the cases, against 20.5\% for a text model retrained with clock words. With one distractor, a multi-object \method{} picks the gazed object more often than a text baseline (47.4\% vs.\ 15.3\%, supplementary Tab.~\suppref{tab:supp_multiobj}{S4}). Second, we propose \textbf{Geometry-Aware Gaze Encoding (\geogaze{})}, whose displacement-alignment term encodes the approach side. Third, we construct \textbf{\dataset{}}, a grasp-centered benchmark that extends HOT3D~\citep{hot3d} with part annotations and object-disjoint three-fold splits for unseen objects.

%% file: sec/2_related.tex
\section{Related work}
\label{sec:related}

\paragraph{HOI motion and grasp generation.}
Recent text-conditioned full-body human--object interaction methods jointly generate human and object motion, guided either by contact, as in HOI-Diff~\citep{hoi-diff}, CG-HOI~\citep{cg-hoi}, and CHOIS~\citep{chois}, or by relational constraints, as in THOR~\citep{thor}.

Hand--object methods emphasize articulated grasps and local contact~\citep{contactgrasp,contactgen,goal,saga,dgrasp,artigrasp,graspxl}. \texthoi{} predicts a text-conditioned surface contact map to guide diffusion and hand refinement~\citep{text2hoi}. DiffH2O~\citep{diffh2o} uses BPS object features~\citep{bps} and separate diffusion models for grasping and manipulation. LatentHOI improves unseen-object generalization by generating global motion and grasp codes with diffusion. A GraspVAE decodes the grasp codes into articulated hands~\citep{latenthoi}. Text2Grasp studies object-part-conditioned static grasps~\citep{text2grasp}. LatentAct instead conditions on an explicit 3D contact point~\citep{latentact}. OpenHOI grounds open-vocabulary affordance regions with a 3D multimodal large language model (LLM) and conditions diffusion on these regions~\citep{openhoi}.

Language provides semantic control but specifies the grasp region only indirectly, and even affordance grounding must infer the region from the wording. Moreover, neither language nor a contact point tells where the person is relative to the object, and this position decides from which side the hand reaches it. \method{} instead grounds recorded gaze rays on object geometry. These rays start at the person's eye and convey both the grasp region and the approach side.

\paragraph{Gaze-conditioned motion generation.} Most gaze-conditioned methods forecast future full-body motion from past motion and gaze. The MoGaze dataset shows that gaze helps predict pick-and-place intent~\citep{mogaze}, and GazeMotion uses gaze to forecast body motion~\citep{gazemotion}. GIMO predicts future motion from past motion, 3D gaze, and the scene, exchanging information between gaze and motion branches~\citep{gimo}, and GazeMoDiff fuses gaze and observed motion with graph attention, then conditions diffusion through cross-attention~\citep{gazemodiff}. These methods predict body motion and model neither hand articulation nor hand--object contact. 

Closest to our setting, GHO-Diffusion synthesizes hand--object interaction from gaze: it represents each frame's gaze by the inverse-distance-weighted feature of the nearest object point, aggregates these features with self-attention, and generates object motion and then hand motion~\citep{gazehoi}. \method{} instead encodes gaze as ray closeness and displacement alignment over fixed BPS correspondences, and conditions the motion on these cues with cross-attention.
The concurrent Gaze2HOI~\citep{ma2026gaze2hoi} introduces the EgoAIM benchmark and forecasts the next second of egocentric interaction on HOT3D-Aria from observed video, gaze, and motion history. \method{} instead synthesizes the full grasp-and-lift motion from 3D gaze rays over object geometry, without text or motion history.

\paragraph{HOI datasets.} ObMan~\citep{obman}, ContactPose~\citep{contactpose}, and OakInk~\citep{oakink} provide static, contact-rich grasps, and HO-3D~\citep{honnotate}, DexYCB~\citep{dexycb}, GRAB~\citep{grab}, H2O~\citep{h2o}, ARCTIC~\citep{arctic}, OakInk2~\citep{oakink2}, and TACO~\citep{taco} provide dynamic 3D hand--object motion, but none of them records gaze.

The GazeHOI dataset of GHO-Diffusion~\citep{gazehoi} records 3D gaze together with hand and object trajectories for task-level HOI synthesis. Its tasks emphasize target selection, rearrangement, and assembly rather than grasped parts, so the dataset is less suited to part-level grasp control. HOT3D~\citep{hot3d} also pairs hand and object poses with eye gaze recorded by Aria glasses. From it we build \dataset{}, with grasp clips, part annotations, and object-disjoint splits for unseen objects.

%% file: sec/3_method.tex
\section{Method}
\label{sec:method}

\subsection{Problem formulation}\label{sec:problem}

\begin{figure*}[t]
\centering
\includegraphics[width=\textwidth]{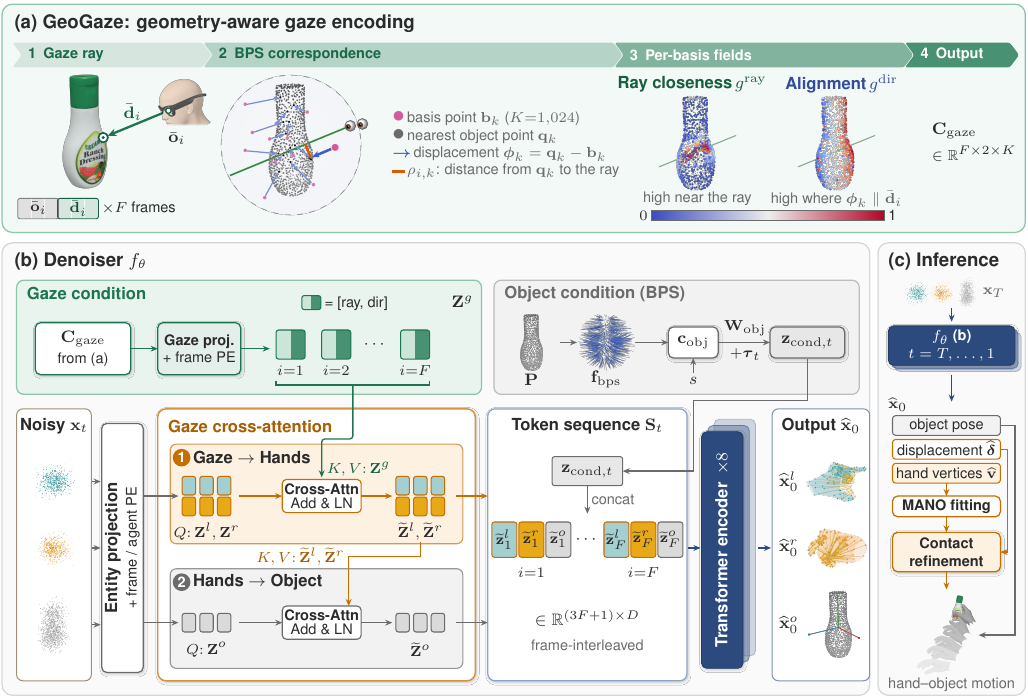}
\caption{\textbf{\method{} overview.} (a)~For one recorded gaze ray, expressed in the initial object frame, \geogaze{} scores each BPS basis point by ray closeness $g^{\mathrm{ray}}$ and displacement alignment $g^{\mathrm{dir}}$, both in $[0,1]$. (b)~In the denoiser $f_\theta$, cross-attention lets the gaze tokens update the hand tokens (1), which then update the object tokens (2). The Transformer predicts $\widehat{\mathbf x}_0$ from $\mathbf z_{\mathrm{cond},t}$ and the hand and object tokens. (c)~Inference: MANO fitting to $\widehat{\mathbf v}$ and contact refinement with $\widehat{\boldsymbol\delta}$ (\cref{sec:inference}).}
\label{fig:method_overview}
\end{figure*}

\Cref{fig:method_overview} summarizes the generation pipeline. We consider offline synthesis of a clip of known length, in which the gaze sequence is temporally aligned with the motion to be generated. \method{} models left-hand, right-hand, and object motion conditioned on the object's normalized geometry and the gaze of the first 30\% of the frames. Let $\mathcal{T}_1^o\in\mathrm{SE}(3)$ be the world-to-object rigid transform of the first frame. This transform is used only to express the gaze and the motion in the initial object frame and is not fed to the denoiser. Only the first-frame object pose is given. The object poses of all later frames are part of the generated motion.

Let $F$ denote the number of motion frames (after padding) and $i\in\{1,\ldots,F\}$ the frame index. At frame $i$, the gaze ray is $\boldsymbol{\gamma}_i=[\mathbf{o}_i,\mathbf{d}_i]\in\mathbb{R}^{6}$, where $\mathbf{o}_i\in\mathbb{R}^3$ is its origin and $\mathbf{d}_i\in\mathbb{R}^3$ is its unit direction. The input gaze sequence is $\boldsymbol{\Gamma}=[\boldsymbol{\gamma}_1,\ldots,\boldsymbol{\gamma}_F]^\top \in\mathbb{R}^{F\times6}$. The normalized object point cloud is denoted by $\mathbf{P}\in\mathbb{R}^{N\times3}$, where $N$ is the number of sampled object points. The uniform scale $s$ normalizes the raw point cloud to a maximum radius of $r_{\mathrm{norm}}$ (\cref{sec:bps}). $\boldsymbol{\mu}$ collects the frame-validity mask and the annotated active-hand flags. Given $\mathbf{P}$, $\boldsymbol{\Gamma}$, $s$, and $\boldsymbol{\mu}$, we model
\begin{equation}
    p_{\theta}(\mathbf{x}_0\mid\mathbf{P},\boldsymbol{\Gamma},s
    ,\boldsymbol{\mu}),
\end{equation}
where $\theta$ denotes the learnable model parameters and $\mathbf{x}_0$ denotes the clean hand--object motion sequence.

For each hand $h\in\{l,r\}$, let $\mathbf{v}_{i,j}^h\in\mathbb{R}^3$ denote the position of the $j$-th sampled MANO surface vertex~\citep{mano} at frame $i$, where $j\in\{1,\ldots,M\}$. We use a fixed subset of $M$ of the 778 MANO vertices, shared across samples. The object point nearest to $\mathbf{v}_{i,j}^h$ at frame $i$ is denoted by $\mathbf{y}_{i,j}^h\in\mathbb{R}^3$, and the corresponding hand-to-object displacement, which encodes the direction and distance to the object as in BimArt~\citep{bimart}, is $\boldsymbol{\delta}_{i,j}^h=\mathbf{y}_{i,j}^h-\mathbf{v}_{i,j}^h$. The per-vertex hand feature~is
\begin{equation}
    \mathbf{u}_{i,j}^h
    = [\mathbf{v}_{i,j}^h,\boldsymbol{\delta}_{i,j}^h]
    \in\mathbb{R}^{6}.
\end{equation}
Concatenating the $M$ vertex features gives the frame-wise hand state:
\begin{equation}
    \mathbf{x}_{0,i}^h
    = [\mathbf{u}_{i,1}^h,\ldots,\mathbf{u}_{i,M}^h]
    \in\mathbb{R}^{6M},
    \qquad h\in\{l,r\}.
\end{equation}

The frame-wise object state is $\mathbf{x}_{0,i}^o=[\mathbf{t}_i^o,\mathbf{r}_i^o]\in\mathbb{R}^{9}$, where $\mathbf{t}_i^o\in\mathbb{R}^3$ and $\mathbf{r}_i^o\in\mathbb{R}^6$ represent the object translation and 6D rotation~\citep{rot6d} relative to its initial pose, respectively.

Over $F$ frames, the clean motion is $\mathbf{x}_0=(\mathbf{x}_0^l,\mathbf{x}_0^r,\mathbf{x}_0^o)$, where $\mathbf{x}_0^h\in\mathbb{R}^{F\times6M}$ for $h\in\{l,r\}$ and $\mathbf{x}_0^o\in\mathbb{R}^{F\times9}$. At diffusion timestep $t$, the denoiser~\citep{ddpm,mdm} reconstructs $\mathbf{x}_0$ from its noisy counterpart $\mathbf{x}_t$:
\begin{equation}
    \widehat{\mathbf{x}}_0
    =f_{\theta}(\mathbf{x}_t,t\mid\mathbf{P},\boldsymbol{\Gamma},s
    ,\boldsymbol{\mu}).
\end{equation}

\subsection{BPS-based object representation}\label{sec:bps}

We center the raw object points $\mathbf p_n^{\mathrm{raw}}$ at their centroid $\mathbf p_{\mathrm c}$ and scale them uniformly, $\mathbf p_n=(\mathbf p_n^{\mathrm{raw}}-\mathbf p_{\mathrm c})/s$, so that $\mathbf P=[\mathbf p_1,\ldots,\mathbf p_N]^\top$ has maximum radius $r_{\mathrm{norm}}<1$ and lies inside the unit ball of the BPS. The gaze is normalized with the same scale $s$, which is also given to the model as the object size. We then encode $\mathbf{P}$ using
$K$ fixed basis points
$\mathbf{B}=\{\mathbf{b}_k\}_{k=1}^{K}$ sampled within the unit
ball~\citep{bps}. For each basis point, we find its nearest point in $\mathbf{P}$ and compute the displacement to it:
\begin{equation}
    \mathbf{q}_k
    =
    \operatorname{NN}_{\mathbf{P}}(\mathbf{b}_k),
    \qquad
    \boldsymbol{\phi}_k
    =
    \mathbf{q}_k-\mathbf{b}_k.
\end{equation}

The object descriptor concatenates these displacements:
\begin{equation}
    \mathbf{f}_{\mathrm{bps}}
    =
    [\boldsymbol{\phi}_1,\ldots,\boldsymbol{\phi}_K]
    \in\mathbb{R}^{3K}.
\end{equation}

\subsection{Geometry-aware gaze encoding (\geogaze{})}

\geogaze{} encodes each gaze ray over the same BPS correspondences used for object representation. Ray closeness, computed from the gaze origin and direction, marks the object points near the ray. Displacement alignment compares the gaze direction with the BPS displacement of every basis point. It does not depend on object size, remains defined when the ray misses the object, and encodes the approach side (supplementary Tab.~\suppref{tab:design}{S10}).

We express each ray in the initial object frame, transforming its origin by $\mathcal T_1^o$ and rotating its direction by the rotation part $\mathbf R_1^o$, and then apply the point-cloud normalization:
\begin{equation}
    \bar{\mathbf{o}}_i
    =\frac{\mathcal{T}_1^o(\mathbf{o}_i)-\mathbf{p}_{\mathrm c}}{s},
    \qquad
    \bar{\mathbf{d}}_i
    =\frac{\mathbf{R}_1^o\mathbf{d}_i}
    {\|\mathbf{R}_1^o\mathbf{d}_i\|_2}.
\end{equation}

\paragraph{Ray closeness.}
For each basis point $\mathbf{b}_k$, let $\rho_{i,k}$ be the distance from its object point $\mathbf{q}_k$ to the forward gaze ray:
\begin{equation}
    \rho_{i,k}
    =
    \min_{\lambda\geq0}
    \left\|
    \mathbf{q}_k-
    (\bar{\mathbf{o}}_i+\lambda\bar{\mathbf{d}}_i)
    \right\|_2.
\end{equation}
A Gaussian kernel maps it to the closeness score
\begin{equation}
    g_{i,k}^{\mathrm{ray}}
    =
    \exp\left(
    -\frac{\rho_{i,k}^{2}}{2\sigma^{2}}
    \right),
\end{equation}
where $\sigma$ controls the spatial bandwidth.

\paragraph{Displacement alignment.}
The displacement $\boldsymbol{\phi}_k=\mathbf{q}_k-\mathbf{b}_k$ points from basis point $\mathbf{b}_k$ toward its corresponding object point $\mathbf{q}_k$. The alignment score is the cosine between $\boldsymbol{\phi}_k$ and the unit gaze direction $\bar{\mathbf d}_i$, rescaled from $[-1,1]$ to $[0,1]$:
\begin{equation}
    g_{i,k}^{\mathrm{dir}}
    =
    \frac{1}{2}\Bigl(1+\frac{\langle\boldsymbol{\phi}_k,\bar{\mathbf d}_i\rangle}{\|\boldsymbol{\phi}_k\|_2}\Bigr).
\end{equation}

For $\alpha\in\{\mathrm{ray},\mathrm{dir}\}$, let $\mathbf g_i^\alpha=[g_{i,1}^\alpha,\ldots,g_{i,K}^\alpha]\in\mathbb R^K$. Stacking both fields over frames gives
\begin{equation}
    \mathbf{C}_{\mathrm{gaze}}
    =
    \big[
    \mathbf{g}_{i}^{\mathrm{ray}},
    \mathbf{g}_{i}^{\mathrm{dir}}
    \big]_{i=1}^{F}
    \in\mathbb{R}^{F\times2\times K}.
\end{equation}

\paragraph{Gaze window.}
Gaze leads the hand: people fixate the object, often where they will grasp it, before the hand arrives~\citep{gaze_1,gaze_2}. The gaze that reveals the grasp intention therefore falls in the approach before contact. In \dataset{}, the hand first comes within 2\,cm of the object at a median of 35\% of the valid frames, and 88\% of the first-30\% gaze frames precede this point (supplementary Sec.~\suppref{sec:supp_clip}{S1.1}). The remaining frames carry a zero gaze token, in training and at inference alike. For a model trained on all-frame gaze, giving only the first 30\% at test time keeps grasp success and costs 2.1 part-accuracy points (supplementary Tab.~\suppref{tab:supp_window}{S1}).

\subsection{Gaze conditioning}

As illustrated in \cref{fig:method_overview}b, the denoiser
conditions the motion on gaze with cross-attention: the gaze tokens update the hand tokens, which then update the object tokens.

Let $D$ denote the latent dimension. The object geometry and scale define
\begin{equation}
    \label{eq:object_condition}
    \mathbf{c}_{\mathrm{obj}}
    =
    [\mathbf{f}_{\mathrm{bps}},s]
    \in\mathbb{R}^{3K+1}.
\end{equation}
At diffusion timestep $t$, we project the object condition with the learned matrix $\mathbf W_{\mathrm{obj}}\in\mathbb{R}^{D\times(3K+1)}$ and add the diffusion timestep embedding $\boldsymbol{\tau}_t$:
\begin{equation}
    \mathbf{z}_{\mathrm{cond},t}
    =
    \mathbf W_{\mathrm{obj}}\mathbf{c}_{\mathrm{obj}}
    +
    \boldsymbol{\tau}_t
    \in\mathbb{R}^{D}.
\end{equation}

Projecting $\mathbf{x}_t$ gives entity tokens $\mathbf{Z}^{l},\mathbf{Z}^{r},\mathbf{Z}^{o}\in\mathbb{R}^{F\times D}$. A shared linear gaze projection maps each frame $i$ of $\mathbf C_{\mathrm{gaze}}$, $[\mathbf g_i^{\mathrm{ray}},\mathbf g_i^{\mathrm{dir}}]\in\mathbb R^{2K}$, to one gaze token that receives a frame positional encoding, so $\mathbf{Z}^{g}\in\mathbb{R}^{F\times D}$. The entity tokens also receive frame and agent-wise positional encodings~\citep{text2hoi}, and validity masks exclude padded tokens from attention.

First, the tokens of each hand $h\in\{l,r\}$ attend to the gaze tokens ($\operatorname{LN}$: layer normalization):
\begin{equation}
    \widetilde{\mathbf{Z}}^{h}
    =
    \operatorname{LN}\!\left(
    \mathbf{Z}^{h}+
    \operatorname{CrossAttn}
    (\mathbf{Z}^{h},\mathbf{Z}^{g})\right).
\end{equation}
This cross-attention lets each hand token retrieve gaze evidence across all gaze frames the model receives.

Let
$\widetilde{\mathbf Z}^{lr}
=[\widetilde{\mathbf Z}^{l},\widetilde{\mathbf Z}^{r}]$
denote the concatenated hand tokens. Second, the object tokens attend to these
gaze-conditioned hand tokens:
\begin{equation}
    \widetilde{\mathbf{Z}}^{o}
    =
    \operatorname{LN}\!\left(
    \mathbf{Z}^{o}+
    \operatorname{CrossAttn}
    \left(
    \mathbf{Z}^{o},
    \widetilde{\mathbf Z}^{lr}
    \right)\right).
\end{equation}

Let $\widetilde{\mathbf{z}}_i^e$ denote the frame-$i$ token of $\widetilde{\mathbf{Z}}^e$, $e\in\{l,r,o\}$. A Transformer encoder maps the token sequence $\mathbf{S}_t$ to $\widehat{\mathbf{x}}_0$, where
\begin{equation}
\begin{aligned}
    \mathbf{S}_t
    = [&\mathbf{z}_{\mathrm{cond},t},
        \widetilde{\mathbf{z}}_1^l,
        \widetilde{\mathbf{z}}_1^r,
        \widetilde{\mathbf{z}}_1^{o},\ldots,\\
       &\widetilde{\mathbf{z}}_F^l,
        \widetilde{\mathbf{z}}_F^r,
        \widetilde{\mathbf{z}}_F^{o}]
    \in\mathbb{R}^{(3F+1)\times D}.
\end{aligned}
\end{equation}

\subsection{Training objective}

The denoiser is trained with a masked reconstruction loss on the clean motion state, $\mathcal{L}=\sum_{e\in\{l,r,o\}}\operatorname{MSE}_{\mathcal{M}_e}(\widehat{\mathbf{x}}_0^e,\mathbf{x}_0^e)$, where $\mathcal{M}_e\subseteq\{1,\ldots,F\}$ contains the valid, unpadded frames of entity $e$. For a hand, $\mathcal{M}_e$ also excludes the frames in which that hand is absent. $\operatorname{MSE}$ averages over these frames and all state channels, including the object translation and rotation.

\subsection{Inference}\label{sec:inference}\leavevmode%
MANO fitting converts the predicted vertices to MANO parameters. From the first predicted contact (hand within 2\,cm of the object) onward, a contact refinement then adjusts the fitted hands for 300 steps with the penetration, displacement, and acceleration terms of BimArt~\citep{bimart}. The displacement term adds the predicted displacement $\widehat{\boldsymbol{\delta}}_{i,j}^h$ to each fitted vertex and penalizes the distance from the result to the object surface. The predicted displacements thus tell the refinement where the hand should touch the object.

%% file: sec/4_experiments.tex
\section{Experiments}
\label{sec:experiments}

\begin{table}[t]
\centering
\caption{\textbf{Comparison on unseen objects} (\%, ID in mm, SimD in cm, three object folds). Each baseline uses its released pipeline. Part: Part Acc., ID: interpenetration depth, Sim: simulation stability. + text prompt: \method{} also receives the baselines' text prompt. Bold: best, underline: second best.}
\label{tab:quantitative_results}
\small
\setlength{\tabcolsep}{0pt}
\begin{tabular*}{\columnwidth}{@{\extracolsep{\fill}}lcccccc@{}}
\toprule
& \footnotesize Success & \multicolumn{2}{c}{\footnotesize Part grounding} & \multicolumn{3}{c@{}}{\footnotesize Physics}\\
\cmidrule(lr{0.6em}){2-2}\cmidrule(lr{0.6em}){3-4}\cmidrule(l{0.6em}){5-7}
Method & GSR\,$\uparrow$ & Part\,$\uparrow$ & PCP\,$\uparrow$ & ID\,$\downarrow$ & Sim\,$\uparrow$ & SimD\,$\downarrow$\\
\midrule
\texthoi{} & 35.3 & 36.8 & 36.6 & 12.0 & 44.0 & 5.09\\
DiffH2O & 17.9 & 31.7 & 31.0 & \phantom{0}8.9 & 27.8 & 7.09\\
LatentHOI & 46.6 & 43.6 & 42.5 & \phantom{0}8.4 & 40.7 & 5.54\\
OpenHOI & 25.8 & 31.5 & 31.0 & 11.9 & 31.7 & 6.16\\
\midrule
\method{} (Ours) & \textbf{66.3} & \underline{49.3} & \underline{50.1} & \phantom{0}\underline{6.7} & \textbf{74.5} & \textbf{2.04}\\
\hspace{0.7em}+ text prompt & \underline{66.0} & \textbf{51.6} & \textbf{52.3} & \phantom{0}\textbf{6.6} & \underline{73.9} & \underline{2.06}\\
\bottomrule
\end{tabular*}
\end{table}

\begin{figure}[t]
\centering
\includegraphics[width=\columnwidth]{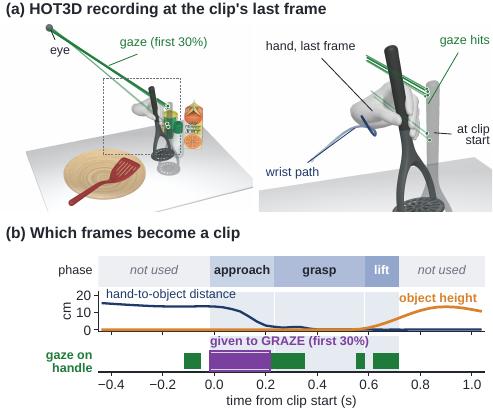}
\caption{\textbf{Grasp-clip construction from a HOT3D recording.} (a)~Last frame of a clip: the gaze rays of the first 30\% of the frames hit the handle of the potato masher (right: the dashed box enlarged). (b)~The clip spans the approach, grasp, and initial lift. The purple box marks the frames whose gaze is given to \method{}.}
\label{fig:clip_construction_example}
\end{figure}

\subsection{Benchmark and protocol}
\begin{figure*}[t]
\centering
\includegraphics[width=\textwidth]{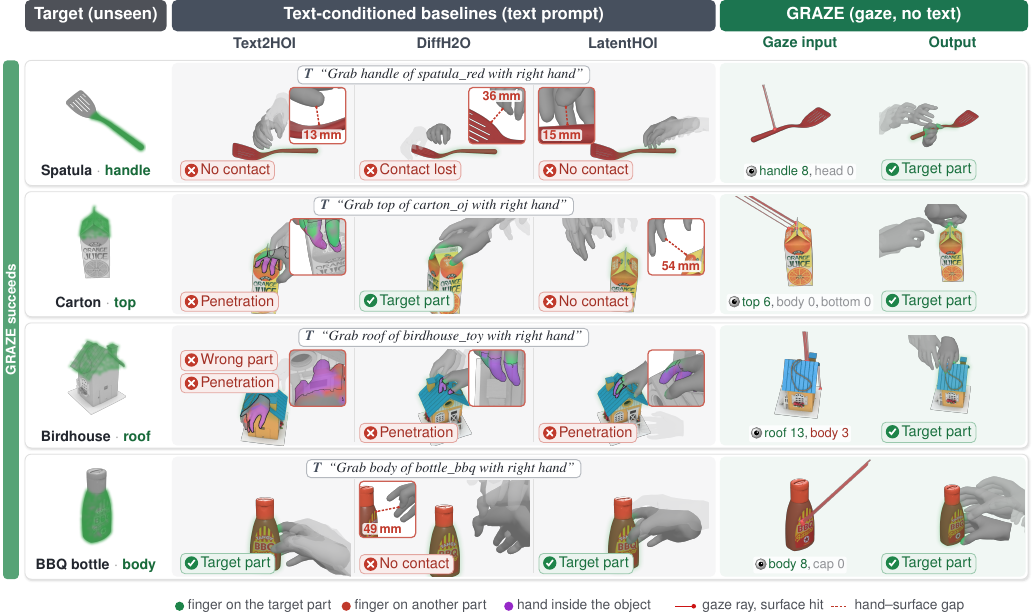}
\caption{\textbf{Generated grasps on unseen objects} (model outputs on test clips), one target part per row (T: text prompt, OpenHOI omitted). Gaze input: rays of the first 30\% of frames, with hits counted per part. Each output is labeled target part, wrong part, no contact, contact lost, or penetration deeper than 10\,mm (definitions in supplementary Sec.~\suppref{sec:supp_failure}{S6}). Close-ups: gap (dashed) or penetrating hand (purple). Fingers on the target part are green, others red. Rows are selected successes of \method{}. Failures are in supplementary Figs.~\suppref{fig:failure_modes}{S6} and~\suppref{fig:gaze_failures}{S7}.}
\label{fig:qualitative_results}
\end{figure*}

We construct \dataset{} from the synchronized hand, object, and gaze recordings of HOT3D~\citep{hot3d} (\cref{fig:clip_construction_example}). It contains 912 grasp clips of 31 objects. Each clip is labeled with the grasped part and runs from the onset of the hand approach until shortly after the object is lifted (median final lift 11.1\,cm, supplementary Sec.~\suppref{sec:supp_clip}{S1.1}). The part labels use 16 names (\eg, handle, rim, body).

Unless noted, experiments use object-disjoint three-fold cross-validation. The 31 objects form three groups, and each fold trains on two groups and tests on the third. Supplementary Sec.~S1 details the construction, annotation, and splits, and shows the object--part distribution (Fig.~\suppref{fig:dataset_distribution}{S1}).%

\subsection{Experimental setup}\label{sec:setup}

\paragraph{Baselines.}
We compare \method{} with four text-conditioned HOI generation methods: \texthoi{}~\citep{text2hoi}, DiffH2O~\citep{diffh2o}, LatentHOI~\citep{latenthoi}, and OpenHOI~\citep{openhoi}. We train all baselines on each fold with their authors' settings, including their refinement (\texthoi{} and OpenHOI train a separate refiner). For OpenHOI, as in its authors' recipe for a new dataset, we train the contact estimator and the generator and use the released affordance-tuned multimodal LLM. DiffH2O and LatentHOI do not take the clip length as a condition, so we set their output length to the clip length (supplementary Sec.~\suppref{sec:supp_crossmodal}{S1.3}). GHO-Diffusion~\citep{gazehoi} releases only a data toolkit, so we evaluate GHO-style gaze features on our backbone instead (\cref{sec:exp_convey}, supplementary Tab.~\suppref{tab:convey}{S9}).

The text baselines receive object geometry and a prompt naming the target object and part. \method{} receives the same geometry and gaze instead. All methods use the same test clips and are evaluated on the annotated active hand. The inputs differ in one more way: \method{} and the same-backbone controls of \cref{sec:exp_convey} are also given the object's first-frame pose, which the baselines do not use. This pose serves only to transform the gaze into the object frame and to place the generated motion in the scene. 
Supplementary Tables~\suppref{tab:supp_shared_pp}{S6} and~\suppref{tab:supp_gazehoi_pp}{S8} also apply our refinement, with nearest-point displacements, to the baselines.

\paragraph{Metrics.}
\emph{Grasp success:} the Grasp Success Rate (GSR), our primary metric, counts a grasp as successful when the active hand touches the object (within 5\,mm) in each of the final five valid frames and, at the final frame, penetrates it by at most 10\,mm and has lifted it by at least 5\,mm. \emph{Part grounding:} Part Contact Precision (PCP), adapted from the contact precision of NL2Contact~\citep{nl2contact}, is the fraction of object points touched in the final frame that lie on the target part. Part Accuracy (Part Acc.), adapted from Text2Grasp~\citep{text2grasp}, is one when the target part holds the most touched points. Both are zero without contact. \emph{Physical plausibility:} Interpenetration Depth (ID) measures how deep the hand enters the object~\citep{diffh2o,latenthoi}, averaged over the frames in which it penetrates. The final-contact rate (Contact) is the proportion of samples that meet the contact condition of GSR. For grasp stability, we fix the final hand and simulate the object under gravity~\citep{obman,contactgrasp}. Sim is the proportion of samples whose object moves less than 2\,cm, and SimD is the mean object displacement (supplementary Sec.~\suppref{sec:supp_sim}{S4}). \Cref{tab:quantitative_results} averages ten samples per test clip, and differences are tested with clip-level bootstrap 95\% confidence intervals (CIs, n.s. if the interval includes zero).

\paragraph{Implementation details.}

Clips are padded to $F=100$ frames. Each object is represented by $N=1{,}024$ points and $K=1{,}024$ basis points ($r_{\mathrm{norm}}=0.85$), and each hand by $M=100$ MANO vertices. The 8-layer diffusion Transformer~\citep{transformer} ($D=512$) uses a 1,000-step cosine diffusion schedule~\citep{iddpm}, sampled with all 1,000 steps, and is trained with AdamW~\citep{adamw} for 4,000 epochs, with a 10\% gaze-condition dropout for training stability. The ray-closeness bandwidth is $\sigma=0.15$ (supplementary Fig.~\suppref{fig:closeness_bandwidth}{S4}).

\subsection{Comparison on unseen objects}
\label{sec:exp_main}

In \cref{tab:quantitative_results}, without being told the part, \method{} grasps the target part more often than every baseline (Part Acc.\ 49.3\% vs.\ 43.6\% for the strongest, LatentHOI). Against every baseline, \method{} also has a higher PCP (50.1\%), a higher GSR on every fold (mean 66.3\%, per fold in supplementary Tab.~\suppref{tab:supp_perfold}{S7}), shallower penetration (6.7\,mm), and more stable grasps in simulation (74.5\% vs.\ at most 44.0\%, all significant). On subject-disjoint folds, where every test object is seen in training, \method{} keeps the highest GSR while \texthoi{} has the highest Part Acc.\ (supplementary Tab.~\suppref{tab:supp_main_table1_sd}{S13}).

\begin{table}[t]
\centering
\caption{\textbf{Zero-shot transfer to GazeHOI}~\citep{gazehoi}. HOT3D object-fold models, no fine-tuning, 939 clips, one sample per clip (\%, ID in mm, SimD in cm). Bold: best.}
\label{tab:gazehoi}
\small
\setlength{\tabcolsep}{0pt}
\begin{tabular*}{\columnwidth}{@{\extracolsep{\fill}}lccccc@{}}
\toprule
Method & GSR\,$\uparrow$ & Sim\,$\uparrow$ & SimD\,$\downarrow$ & Contact\,$\uparrow$ & ID\,$\downarrow$\\
\midrule
\texthoi{} & 18.4 & 27.1 & 7.50 & 46.8 & 11.2\\
DiffH2O & 11.1 & 22.7 & 8.00 & 35.3 & 12.8\\
LatentHOI & 33.8 & 48.6 & 4.80 & 72.6 & 11.6\\
OpenHOI & 16.8 & 34.0 & 6.32 & 51.8 & 13.3\\
\midrule
\method{} (Ours) & \textbf{45.3} & \textbf{67.2} & \textbf{2.97} & \textbf{92.2} & \textbf{10.3}\\
\bottomrule
\end{tabular*}
\end{table}

\begin{figure*}[t]
\centering
\includegraphics[width=\textwidth]{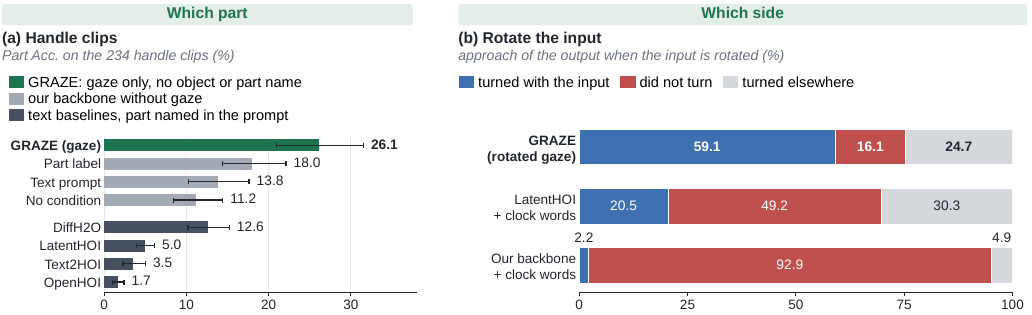}
\caption{\textbf{What gaze conveys.} (a)~Which part: Part Acc.\ on the 234 handle clips. Error bars: 95\% CIs. (b)~Which side: inputs rotated about the object's vertical axis by 90$^\circ$, 180$^\circ$, or 270$^\circ$ (gaze for \method{}, clock word for the text models). An output turned with the input when its final wrist moved around the object by the rotation ($\pm$45$^\circ$), did not turn when it stayed within 45$^\circ$ of the unrotated output, and turned elsewhere otherwise.}
\label{fig:convey}\label{fig:control}
\end{figure*}

The GSR lead remains when every baseline also receives our refinement with nearest-point displacements (supplementary Tab.~\suppref{tab:supp_shared_pp}{S6}). \Cref{fig:qualitative_results} shows examples. Supplementary Sec.~\suppref{sec:supp_crossmodal}{S1.3} analyzes where each baseline falls short.

Gaze accounts for part of this lead: on our backbone, it raises GSR from 59.0\% without any condition to 66.3\% (significant, supplementary Tab.~\suppref{tab:convey}{S9}). The rest comes from the displacements predicted for refinement (\cref{sec:inference}): if both use nearest-point displacements, LatentHOI leads by 8.3 GSR points (supplementary Tab.~\suppref{tab:supp_shared_pp}{S6}).

\paragraph{Zero-shot transfer.}We apply the HOT3D object-fold models, without fine-tuning, to 939 clips of GazeHOI~\citep{gazehoi}, a separate gaze dataset with different objects (protocol in supplementary Sec.~\suppref{sec:supp_gazehoi}{S1.4}). \method{} leads the strongest baseline, LatentHOI, by 11.5 GSR points (\cref{tab:gazehoi}) and keeps contact through the final five frames in 92.2\% of the samples. \method{} also penetrates least and gives the most stable grasps in simulation.

\subsection{What does gaze convey?}
\label{sec:exp_convey}

We fix the backbone and the training recipe and vary only the condition: no condition, the target part label, the text prompt (CLIP-encoded~\citep{clip}), and other encodings of the same gaze (supplementary Tab.~\suppref{tab:convey}{S9}). Differences are significant unless marked n.s.

\paragraph{Which part.}Gaze raises Part Acc.\ by 5.6 and PCP by 6.5 points over no condition and is not significantly different from the ground-truth part label. The gain is clearest on handles: a handle sticks out to one side, so a gaze toward that side indicates it (Part Acc.\ 26.1\% vs.\ at most 18.0\% without gaze, \cref{fig:convey}a). Caps, rims, tops, and lids are harder: people also look at the top of an upright object in 84\% of body grasps, so such gaze does not single out these parts. On these parts the text prompt is stronger (34.7\% vs.\ 19.6\%), and over all parts it is slightly ahead (52.2\% vs.\ 49.3\%, supplementary Sec.~\suppref{sec:supp_rotation}{S5.2}).

\paragraph{Which side.}Gaze also conveys the approach side, which the prompt and the part label omit. We rotate the recorded gaze around the object by 90$^\circ$, 180$^\circ$, and 270$^\circ$, as if the person looked from another side. The final wrist of \method{} turns with the rotation (within 45$^\circ$) for 59.1\% of the rotated inputs, whereas LatentHOI retrained with clock words follows a changed word in only 20.5\% (chance: 25\%, \cref{fig:control}b, supplementary Sec.~\suppref{sec:supp_rotation}{S5.2} and Fig.~\suppref{fig:rot_example}{S5}).  The eye position alone carries much of the benefit: a model given only the eye-to-centroid ray (Head-only, supplementary Tab.~\suppref{tab:convey}{S9}) beats no condition (GSR $+4.1$, Part Acc.\ $+4.3$). \method{} leads it in GSR ($+3.2$) but not in Part Acc.\ ($+1.3$, n.s.). The gaze direction matters on the 11\% of clips whose gaze deviates from this ray by at least 10$^\circ$: there, \method{} exceeds Head-only by 19 GSR points. The raw ray and a GHO-style feature~\citep{gazehoi} follow the rotation less often than \method{} (50.8\% and 53.7\% vs.\ 59.1\%).

\paragraph{Which object.}A multi-object variant of \method{} receives the gaze and every candidate object. With one or two distractors, it contacts only the unseen target in 47.4\% and 34.6\% of the scenes, against 15.3\% and 4.0\% for \texthoi{} extended to several objects and given the text prompt (supplementary Sec.~\suppref{sec:supp_multiobj}{S2.3} and Tab.~\suppref{tab:supp_multiobj}{S4}).

\paragraph{Gaze and text.}When a prompt is also available, the two inputs complement each other (Tab.~\ref{tab:quantitative_results} and supplementary Tab.~\suppref{tab:convey}{S9}). Adding gaze to the prompt keeps its GSR and Part Acc.\ (n.s.) and lets the approach follow a rotated gaze in 57.4\% of the rotated inputs (gaze alone: 59.1\%, n.s.), which a prompt alone cannot do. Adding the prompt to gaze raises PCP from 50.1\% to 52.3\%, the highest of all methods, and Part Acc.\ from 49.3\% to 51.6\% (n.s.).

\subsection{Robustness to gaze noise}\label{sec:exp_control}
Real eye tracking is imperfect, so we corrupt the recorded gaze at test time, without retraining. We either add random angular noise to every gaze direction (RMS error of 5$^\circ$ or 10$^\circ$), or drop each gaze frame with probability 0.2 or 0.5, which zeroes its gaze token. In every case, GSR and Part Acc.\ change by at most 1.1 points (n.s., supplementary Tab.~\suppref{tab:supp_gaze_noise}{S11}).

%% file: sec/5_conclusion.tex
\section{Conclusion}
\label{sec:conclusion}

 We presented \method{}, which generates grasp-and-lift motion from the first 30\% of the recorded gaze instead of object and part names, together with \dataset{}, a benchmark with object-disjoint folds. On unseen objects, \method{} grasped the target part more often than the strongest text baseline and achieved the highest grasp success. Without fine-tuning, it also achieved the highest grasp success on the GazeHOI dataset. Unlike the text models, it turned its approach with rotated gaze, and a multi-object variant selected the gazed object among distractors more often.

Gaze also has limits. People do not always look at the part they grasp, \eg, at the top of an object grasped by its body. Adding a text prompt to gaze raised PCP, while gaze still set the approach side. Future work will fuse gaze and language more tightly when a prompt is available, generate motion online from a live eye tracker, and extend to gaze-driven pick-and-place in whole scenes.  

%% file: sec/6_llm_statement.tex
\section*{LLM Usage Statement}
We used Claude (Anthropic), a large language model, through its agentic
coding tool Claude Code. (i)~\emph{Writing}: polishing the wording, grammar, and
consistency of author-written text in the paper and the supplementary material,
without introducing new claims, and obtaining reviewer-style feedback on the manuscript. (ii)~\emph{Baselines}: writing and debugging the
scripts that train and evaluate \texthoi{}, DiffH2O, LatentHOI, and OpenHOI on our benchmark,
configuring them to follow the official releases where available, and launching and monitoring
these runs. (iii)~\emph{Ablations of our model}: implementing and running ablation
and analysis experiments, including gaze-input variants (\eg, removing, rotating,
or temporally restricting the gaze), with the evaluation and statistical-analysis
scripts. (iv)~\emph{Related work, references, and figures}: searching for related
and concurrent work, checking bibliographic metadata, and producing figures (rendering scripts and layout). The research idea, the method, and the benchmark construction are the
authors' own. The authors reviewed and verified all LLM-assisted text, code, and
results, and take full responsibility for the content of this paper.\par

%% file: sec/X_suppl.tex
\raggedbottom
\renewcommand{\dblfloatpagefraction}{0.8}
\setcounter{section}{0}\setcounter{figure}{0}\setcounter{table}{0}\setcounter{equation}{0}
\maketitlesupplementary

\renewcommand{\thefigure}{S\arabic{figure}}
\renewcommand{\thetable}{S\arabic{table}}
\renewcommand{\theequation}{S\arabic{equation}}
\renewcommand{\thesection}{S\arabic{section}}

\section{HOT3D-Grasp benchmark construction and evaluation protocols}

\begin{figure}[t]
\centering
\includegraphics[width=\columnwidth]{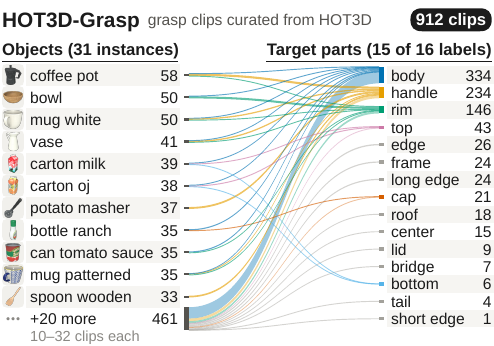}
\caption{\textbf{Object--part distribution of \dataset{}.} Each flow connects an object instance to its target part, named by the surface label used for part-level scoring. Counts denote clips.}
\label{fig:dataset_distribution}
\end{figure}

\Cref{fig:dataset_distribution} summarizes the object and grasp-part coverage of \dataset{}.

\subsection{Clip construction}\label{sec:supp_clip}

\dataset{} is derived from long HOT3D~\citep{hot3d} scenario recordings with
synchronized hand, object, and gaze annotations. We manually segment each grasp
from the onset of the hand approach until shortly after the object leaves the table (final lift: median 11.1\,cm, at least 5\,cm in 74\% of the clips). Each clip retains the available left- and
right-hand MANO~\citep{mano} motion, the target-object pose and geometry, and the
frame-level gaze ray. We use a fixed input length of $F=100$ frames, as clips have at most 92 frames (37 on average).
\ifwithsupp\Cref{fig:clip_construction_example}\else Figure~3 of the main paper\fi{} illustrates this temporal
segmentation with an example clip cut from one HOT3D recording, together with the gaze input of its first 30\% of frames. This window mostly covers the approach: in the ground truth, the hand first comes within 2\,cm of the object after a median of 35\% of the valid frames, and 88\% of the first-30\% gaze frames precede it.

Three annotators verify the boundaries of each clip, assign its object and active hand, name a grasp part when the
dominant contact region forms a semantically distinguishable object part, and
segment that part directly on the object surface.

\paragraph{Gaze window.}To test whether this window discards useful gaze, we train a model on the gaze of all frames and vary the window at inference (\cref{tab:supp_window}). The before-contact window keeps the frames up to the first ground-truth contact (a hand joint within 2\,cm). No window changes GSR significantly relative to the first 30\%. Only the all-frames window raises Part Acc.\ and PCP significantly, by 2.1 and 1.9 points (48.6\% vs.\ 46.5\% and 49.7\% vs.\ 47.8\%).

\begin{table}[t]
\centering
\caption{\textbf{Gaze window at inference.} A separate model, trained on the gaze of all frames rather than on the first 30\% as for the main model, receives a different window at test time (three object folds, \%). $^\ast$: significantly different from the first 30\% (clip-paired bootstrap 95\% CI). Bold: best value per column.}
\label{tab:supp_window}
\small
\setlength{\tabcolsep}{0pt}
\begin{tabular*}{\columnwidth}{@{\extracolsep{\fill}}lccc@{}}
\toprule
Gaze window & GSR\,$\uparrow$ & Part Acc.\,$\uparrow$ & PCP\,$\uparrow$\\
\midrule
First 10\% & 61.9 & 47.9 & 48.6\\
First 20\% & \textbf{64.5} & 47.4 & 48.7\\
First 30\% & 63.8 & 46.5 & 47.8\\
First 50\% & 63.3 & 46.5 & 48.1\\
Before contact & 63.2 & 46.1 & 47.9\\
All frames & 63.8 & \textbf{48.6}$^\ast$ & \textbf{49.7}$^\ast$\\
\bottomrule
\end{tabular*}
\end{table}

\begin{figure*}[t]
\centering
\includegraphics[width=0.9717\textwidth]{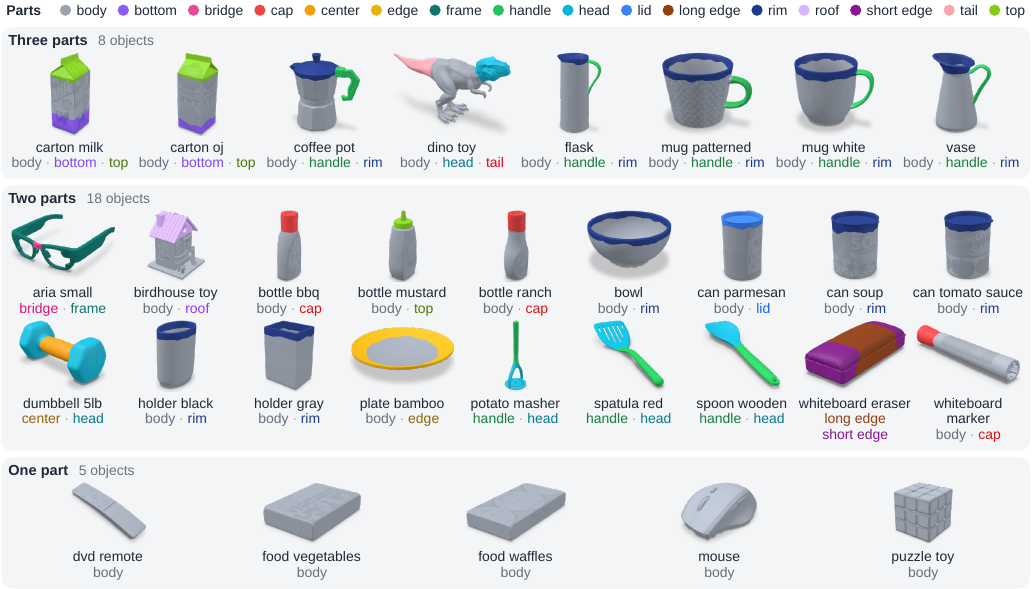}
\caption{\textbf{Semantic surface segmentation of all benchmark objects.}
Every one of the 1,024 sampled object points is assigned to one of at most
three canonical surface parts. Objects are grouped by annotation complexity:
three parts (top), two parts (middle), and one part (bottom). The same color
denotes the same canonical part across all objects. Part regions follow the recorded grasps. For the eraser, ``long edge'' grasps hold its middle across the long sides and ``short edge'' grasps hold one end.}
\label{fig:part_segmentation_inventory}
\end{figure*}

\paragraph{Part annotation.}Each object surface has at most three semantic parts, and the segment of the grasped part is the target region for part-level evaluation. The prompt names the part, the object, and the active hand (\eg, ``Grab handle of spatula\_red with right hand.'').
\Cref{fig:part_segmentation_inventory} visualizes the complete
surface-segmentation inventory. The 31 objects comprise 8 three-part,
18 two-part, and 5 one-part objects.

Of the 33 HOT3D objects, we discard the cellphone and the keyboard, which have fewer than 1,024 surface points, and we drop clips with invalid contact annotations. The
resulting benchmark contains 912 clips, 31 objects, and 16 surface-part labels, 15 of which are grasped in at least one clip. The exception is \textit{head}: the five objects with a head part (masher, spatula, spoon, dumbbell, and dino toy) are always held by another part.

\dataset{} introduces no new human recordings. All recordings and their hand, object, and gaze annotations originate from HOT3D and are used under the HOT3D dataset license, while the clip boundaries and part labels are ours.
Our release provides derived clip indices and annotations and requires users to
obtain the underlying HOT3D data and MANO assets under their respective
licenses.

\subsection{Object- and subject-disjoint folds}
\begin{figure*}[t]
\centering
\includegraphics[width=\textwidth]{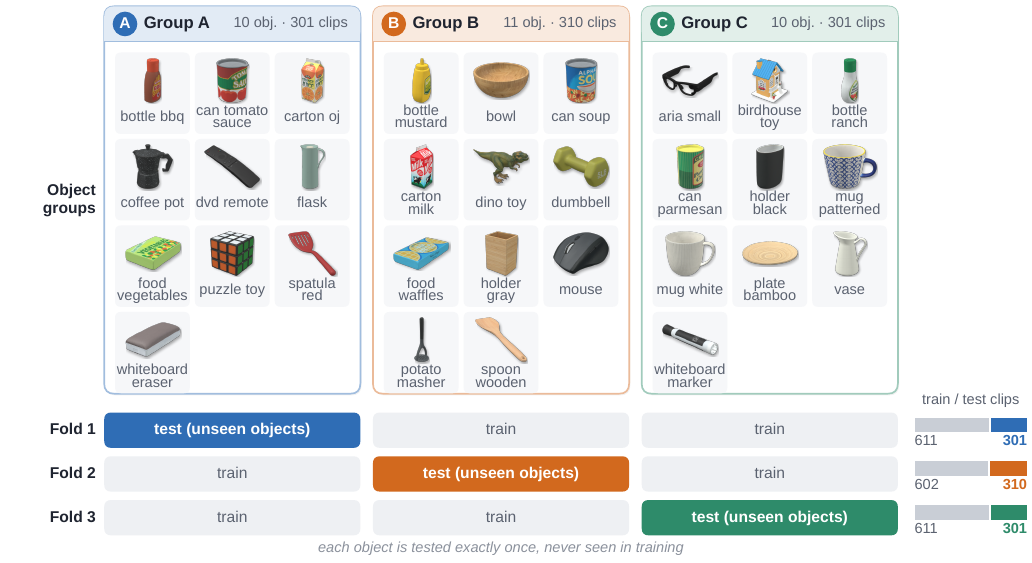}
\caption{\textbf{Object-disjoint three-fold cross-validation.} The 31
objects form three disjoint groups, shown as object cards with one color per group. In fold $k$ (rows), the group
in its color is held out entirely as unseen test objects and the other two groups are used for
training. Bars give the train and test clip counts of each fold.}
\label{fig:threefold_protocol}
\end{figure*}

\leavevmode%
The main comparison uses object-disjoint three-fold cross-validation
(\cref{fig:threefold_protocol}). The 31 objects are partitioned into three
disjoint groups of 10, 11, and 10 objects. Fold $k$ trains on every clip of the
other two groups and tests on every clip of group $k$, so the three test sets
partition the 912 clips and each object is evaluated exactly once, as an object
never seen in training. Every method is trained once per fold with its standard
recipe and evaluated with ten samples per test clip. We report the mean over the
three folds. The subject-disjoint variant instead partitions the nine recording
subjects into three groups of three and tests on the clips of the held-out
subjects. It therefore measures generalization to unseen participants: the test objects
also appear in training, and only one or two object--part combinations per fold are absent from it. Exact sizes are listed in \cref{tab:split_statistics}.

\begin{table}[t]
\centering
\caption{\textbf{Benchmark split statistics.} All protocols reorganize the same 912 clips. The three object test sets partition the benchmark, as do the three subject test sets.}
\label{tab:split_statistics}
\small
\setlength{\tabcolsep}{3pt}
\begin{tabular*}{\columnwidth}{@{\extracolsep{\fill}}lccc@{}}
\toprule
Protocol & \shortstack{Train\\clips} & \shortstack{Test\\clips}
& \shortstack{Objects\\(train/test)}\\
\midrule
Object fold 1 & 611 & 301 & 21 / 10 (unseen)\\
Object fold 2 & 602 & 310 & 20 / 11 (unseen)\\
Object fold 3 & 611 & 301 & 21 / 10 (unseen)\\
Subject fold 1 & 627 & 285 & 31 / 31\\
Subject fold 2 & 603 & 309 & 31 / 30\\
Subject fold 3 & 594 & 318 & 31 / 31\\
\bottomrule
\end{tabular*}
\end{table}

\subsection{Cross-modal comparison and controls}\label{sec:supp_crossmodal}

The comparison between gaze and text is intentionally asymmetric in input modality but controlled in evaluation data (\cref{tab:conditioning_inputs}). Text-conditioned methods receive object geometry and an explicit prompt containing the object and target part. The active hand (left, right, or both) is given to every method, as a per-hand flag for \texthoi{}~\citep{text2hoi}, OpenHOI~\citep{openhoi}, and \method{}, and through the prompt for DiffH2O~\citep{diffh2o} and LatentHOI~\citep{latenthoi}. \method{} receives the same object geometry and the recorded gaze of the first 30\% of the frames, without object- or target-part text. Only \method{} and its same-backbone controls (\cref{tab:convey,tab:design}) also receive the first-frame object pose, which serves only to express the gaze in the object frame and to place the generated motion in the scene. All methods are evaluated on the same clips and object geometries. The experiment tests whether gaze, without object or part text, conveys the part that a prompt names and the approach side that it leaves open.

\begin{table}[t]
\centering
\caption{\textbf{Conditioning inputs and geometry encoders.} Geometry
denotes the object-shape encoder. Hand: how the active hand is given to the method. Length: the method generates exactly the clip length. Init.\ pose is the first-frame object pose.}
\label{tab:conditioning_inputs}
\footnotesize
\setlength{\tabcolsep}{0pt}
\begin{tabular*}{\columnwidth}{@{\extracolsep{\fill}}lcccccc@{}}
\toprule
Method & Geometry & Text & Hand & Gaze & Length & Init.\ pose\\
\midrule
\texthoi{} & PointNet & \yesmark & flag & \nomark & \yesmark & \nomark\\
DiffH2O & BPS & \yesmark & prompt & \nomark & \yesmark & \nomark\\
LatentHOI & BPS & \yesmark & prompt & \nomark & \nomark & \nomark\\
OpenHOI & PointNet & \yesmark & flag & \nomark & \yesmark & \nomark\\
\method{} & BPS & \nomark & flag & \yesmark & \yesmark & \yesmark\\
\bottomrule
\end{tabular*}
\end{table}

DiffH2O and LatentHOI are originally formulated with separate grasping and
manipulation stages. For DiffH2O, we train the authors' full model in its one-stage configuration with short text prompts (its ``simple annotations'') on each fold and refer to this model as DiffH2O throughout. For LatentHOI, we train its complete model with the released code on each fold. Neither method takes the clip length as an explicit condition: DiffH2O generates a clip-length motion with its official zero mask, and the fixed 96-frame output of LatentHOI is resampled in time to the clip length, so that each is judged on the end state of its own motion. We introduce no further adaptation of either two-stage method. In contrast, Text2HOI has no explicit grasp--manipulation stage
separation, so we train its full model directly on our benchmark. We convert
the final motions from all methods to a common hand--object representation and
compute all reported comparison metrics with the same evaluator, thresholds,
frame masks, and object meshes.
OpenHOI~\citep{openhoi} adds an object affordance map, predicted by a multimodal large language model (LLM) from the prompt, to the \texthoi{} contact estimator and generator. We compute the affordance maps with the released affordance-tuned multimodal LLM and, following the authors' recipe for a new HOI dataset, train the contact estimator and the generator on each fold. The maps separate parts on only 4 of the 21 objects whose clips grasp two or more different parts. On the other 17 objects they barely depend on the part name (median correlation 0.99 between maps for different part names). On most objects, OpenHOI thus behaves like \texthoi{} with a weak part cue, consistent with its lower scores in Tab.~\suppref{tab:quantitative_results}{1} of the main paper. To rule out a reproduction error, we follow the released configuration and train the generator for as many epochs as the \texthoi{} generator on which it is built. On every fold it reaches a training loss similar to that of \texthoi{}, so its lower scores do not come from failed training.

Two measurements show where the baselines fall short in Tab.~\suppref{tab:quantitative_results}{1} of the main paper. First, the baselines mostly fail before the grasp is complete. The active hand is not in contact through the final five frames in 30.3\% of the samples of \texthoi{}, 55.3\% of DiffH2O, 31.5\% of LatentHOI, and 45.8\% of OpenHOI, against 1.9\% for \method{}, whose GSR failures are mostly final-frame penetration above 10\,mm (26.7\% of the samples, while \cref{fig:failure_modes} shows a smaller penetration share because its taxonomy assigns wrong-part grasps before penetration). Second, the text-conditioned baselines ground the named part far better on objects seen in training. From the subject-disjoint folds (\cref{tab:supp_main_table1_sd}), where every test object is seen, to the object folds, Part Acc.\ falls from 71.7\% to 36.8\% for \texthoi{}, from 60.9\% to 31.7\% for DiffH2O, and from 69.7\% to 43.6\% for LatentHOI, against 59.2\% to 49.3\% for \method{}. This is consistent with a part name being learned together with the shapes of the training objects, whereas gaze marks the location on the new object itself.

Every method is evaluated after its own conversion to a hand mesh and its own refinement. \Cref{tab:supp_shared_pp} adds the nearest-point refinement (the 300-step contact refinement of \method{}, \cref{sec:supp_impl}, with displacement targets to the nearest object points instead of the displacements predicted by \method{}) to every baseline.

Variants of the same model (prompt and gaze interventions) use the same sampling noise and are therefore paired by clip. All Follow values and the GazeHOI part-word test (\cref{sec:supp_gazehoi}) use four samples per clip, and the zero-shot GazeHOI comparison one sample per clip, instead of ten.

\subsection{Zero-shot GazeHOI protocol}\label{sec:supp_gazehoi}
We convert GazeHOI~\citep{gazehoi} to the format of \dataset{} and give the same converted data to every method. Each clip runs from 12 frames before the first hand--object contact to 11 frames after the lift onset (at most 99 frames, median 45). The hand parameters are converted to our MANO layer, and the scene and the object meshes are rotated so that their world-up direction matches that of HOT3D. The evaluation set contains 939 clips of 36 objects from 194 recordings: every clip of both GazeHOI splits in which the recorded gaze ray comes within 1\,cm of the object, \ie, the 94 clips of 28 objects of the test split and 845 clips of the training split. No evaluated model is trained on GazeHOI, so both splits are unseen by every method. Each method generates one sample per clip. GazeHOI has no part annotation, and its prompts name the hand but neither the object nor a part (\eg, ``Grab this with right hand.''), so every text baseline receives this prompt (\method{} receives only the gaze and the same hand flag) and Part Acc.\ and PCP are not reported. The three object-fold models of \ifwithsupp\cref{tab:quantitative_results}\else Tab.~\suppref{tab:quantitative_results}{1} of the main paper\fi{} are applied without fine-tuning, and we report the mean over the three models. \method{} uses the gaze of the first 30\% of the frames, as on HOT3D. Adding a part word to this prompt does not help the strongest baseline: on the 94 test-split clips, with four samples per clip, LatentHOI reaches a GSR of 30.5\% with the original prompt, 29.9\% with ``Grab body of this with \ldots'' (n.s.), and 23.0\% with ``Grab handle of this with \ldots'' (significantly lower).

\section{Additional method details}

\subsection{Frames, masking, and diffusion}

All inputs and outputs are expressed in the initial object frame (Sec.~\suppref{sec:problem}{3.1} of the main paper), and padded frames and absent hands are masked in attention and in the loss. The denoiser predicts the clean state $\widehat{\mathbf{x}}_0$ from $\mathbf{x}_t=\sqrt{\bar{\alpha}_t}\mathbf{x}_0+\sqrt{1-\bar{\alpha}_t}\boldsymbol{\epsilon}$, where $\bar{\alpha}_t$ is the cumulative noise schedule at diffusion step $t$ and $\boldsymbol{\epsilon}\sim\mathcal N(\mathbf 0,\mathbf I)$, with one reconstruction loss for the hands and the object.

\subsection{Geometry-aware gaze features}

For every valid frame and BPS index, \geogaze{} retains two separate values: ray closeness (a Gaussian of the distance from the nearest object point $\mathbf q_k$ of the basis point to the forward gaze ray) and displacement alignment (the cosine between the gaze direction and the BPS displacement $\boldsymbol\phi_k$, rescaled to $[0,1]$). Closeness uses the gaze origin and direction and remains nonzero when a noisy ray passes near, rather than through, a small object. Alignment uses the direction only and remains defined without an intersection. It encodes the viewing direction in the object frame, not where the ray lands. The two $K$-dimensional fields of a frame are concatenated before the gaze projection, so each frame contributes one gaze token in which the closeness and alignment entries keep separate projection weights.

\begin{figure}[t]
\centering
\includegraphics[width=\columnwidth]{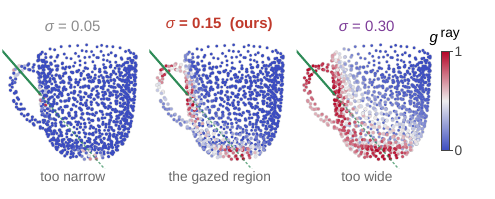}
\caption{\textbf{Choice of the ray-closeness bandwidth $\sigma$.} Closeness on a mug for one gaze ray at three bandwidths (green: the gaze ray, dashed inside the mug). A small $\sigma$ marks almost no point, and a large one spreads over the handle and the body.}
\label{fig:closeness_bandwidth}
\end{figure}

Let $\sigma=0.15$ denote the fixed bandwidth in the normalized object
coordinate system. Since objects are scaled to $r_{\mathrm{norm}}=0.85$, $\sigma$
corresponds to about $17.6\%$ of the maximum object radius. For an object point $\mathbf q_k$ at distance $\rho$ from the forward gaze ray, its relative contribution is $w(\rho)=\exp(-\rho^2/(2\sigma^2))$. \Cref{fig:closeness_bandwidth}
visualizes the closeness field at three bandwidths. Two geometric quantities, measured on the 912 clips independently of any model output, motivate the chosen value. First, the gaze ray often passes near rather than through the object: in 33\% of clips its median distance to the nearest object point before contact exceeds 0.05. A field whose maximum falls below 0.1 carries almost no signal. This happens for 23\% of clips at $\sigma=0.05$ but only 8\% at $\sigma=0.15$. Second, the field should stay part-scale: the median part radius of multi-part objects is 0.38, and at $\sigma=0.15$ the highlighted width (full width at half maximum, FWHM, $2.35\sigma$) is 46\% of a typical part diameter, whereas at $\sigma=0.30$ it reaches 92\% and spans nearly a whole part. We selected this
value as a fixed locality--robustness trade-off and use it for every object,
split, and ablation. It is not changed at test time.

\subsection{Multi-object slots}\label{sec:supp_multiobj}

\paragraph{Multi-object target selection.}Each test scene of the three object folds places the recorded target, an object unseen in training, together with one distractor (two-object setting) or two distractors (three-object setting), drawn from seen and unseen objects. Each setting has one scene per clip (912 scenes). Target Selection Accuracy (TSA) is the proportion of scenes in which the hand contacts the target and no distractor. Distractor Contact Rate (DCR) is the proportion in which it contacts any distractor, so TSA$+$DCR$\leq$100 (contact as defined for multi-object evaluation in \cref{sec:supp_metrics}: at least two active-hand keypoints within 2\,cm). Held is the proportion of scenes whose hand still holds the target in the final five frames.

For $Q$ candidates, the object encoder and \geogaze{} share weights across candidates. Candidate-specific gaze and object features remain in their own slots, while the global object condition is obtained by mean pooling. Two-object training pairs the target with a context object present in the same HOT3D recording and randomly permutes their slot order. The context slot retains its recorded trajectory, and reconstruction losses supervise every object slot. There is no explicit collision loss or target-index embedding. The hand streams attend to all candidate gaze features, so the model must infer which candidate is the target from where the gaze falls relative to each candidate's geometry. The same parameters are used for every slot, allowing zero-shot three-candidate evaluation without new learned parameters.

\paragraph{\texthoi{}$^\ast$ adaptation.}
The original object input becomes slot 1,
and an identically encoded candidate is concatenated as slot 2. Each slot
receives its own PointNet geometry, scale, centroid, trajectory token, and
slot-aligned contact-map embedding. The target slot receives the prompt-derived
contact map and the context slot a zero map. As with \method{}, the two-slot model is evaluated with a
third slot that reuses the slot weights, without fine-tuning.
On the three object folds, \texthoi{}$^\ast$ is trained for about 9,000 epochs and the \method{} multi-object model for 4,000 epochs, so the gaze model receives fewer updates.

\paragraph{Multi-object results.}We compare with \texthoi{} only: its point-cloud object encoder and contact estimator take each additional candidate as one more object slot with minimal changes to the architecture (denoted \texthoi{}$^\ast$), whereas DiffH2O and LatentHOI condition on a single object and would need a structural redesign. \Cref{tab:supp_multiobj} reports one trained model per method and fold, averaged over the three folds. \method{} receives the first 30\% of the gaze frames. \method{} uses its contact refinement, applied to the union of the candidate surfaces. For a fair comparison we also apply the nearest-point refinement of \cref{tab:supp_shared_pp}, likewise on the union of the candidate surfaces, to \texthoi{}$^\ast$. The refinement leaves the scores of \texthoi{}$^\ast$ almost unchanged (within 0.7 points), because it moves only the hands and \texthoi{}$^\ast$ already contacts a distractor in most scenes. \method{} more often contacts the target alone than \texthoi{}$^\ast$ does, touches a distractor less often, and holds the target at the end about twice (two objects) and four times (three objects) as often, but it stays far below the recorded motion. \texthoi{}$^\ast$ moves its distractors by more than 1\,cm in 98--99\% of the two-object scenes across the three folds (\method{}: 26--29\%), so its DCR partly reflects the distractor motion it generates.

\begin{table}[t]
\centering
\caption{\textbf{Multi-object target selection} averaged over the three object folds with two or three candidate objects (\%). $^\ast$: \texthoi{} extended to object slots. + nearest-pt.: with the nearest-point refinement of \cref{tab:supp_shared_pp}. Bold: best generated result.}
\label{tab:supp_multiobj}
\small
\setlength{\tabcolsep}{0pt}
\begin{tabular*}{\columnwidth}{@{\extracolsep{\fill}}lcccccc@{}}
\toprule
& \multicolumn{3}{c}{2 objects} & \multicolumn{3}{c}{3 objects}\\
\cmidrule(lr{.5em}){2-4}\cmidrule(l{.5em}){5-7}
Method & TSA\,$\uparrow$ & DCR\,$\downarrow$ & Held\,$\uparrow$ & TSA\,$\uparrow$ & DCR\,$\downarrow$ & Held\,$\uparrow$\\
\midrule
\texthoi{}$^\ast$ & 15.3 & 79.6 & 19.3 & \phantom{0}4.0 & 93.6 & \phantom{0}7.3\\
\quad + nearest-pt. & 15.0 & 80.3 & 19.7 & \phantom{0}3.9 & 94.0 & \phantom{0}7.5\\
\method{} (Ours) & \textbf{47.4} & \textbf{24.1} & \textbf{42.3} & \textbf{34.6} & \textbf{40.0} & \textbf{30.9}\\
\midrule
Recorded (GT) & 95.6 & \phantom{0}4.4 & 99.9 & 93.3 & \phantom{0}6.7 & 99.7\\
\bottomrule
\end{tabular*}
\end{table}

\section{Implementation and reproducibility}\label{sec:supp_impl}

Because the network predicts vertices rather than MANO parameters, we fit MANO pose, translation, and shape to the $M=100$ predicted vertices of each hand using vertex matching with weak pose, shape, and temporal regularization. We then apply a deterministic contact refinement. Neither stage is learned, and neither uses ground-truth hand pose or ground-truth contact.

MANO fitting uses 80 Adam iterations with learning rate 0.05, pose
and shape regularization $10^{-4}$, and pose- and translation-temporal
regularization $10^{-3}$.
The contact refinement then optimizes the fitted mesh for 300 AdamW steps at learning rate $5\times10^{-4}$ with the BimArt~\citep{bimart} penetration, displacement, and acceleration terms weighted 10, 100, and 1,000. A frame counts as in contact when the predicted hand comes within 2\,cm of the object. The refinement runs from the first predicted contact to the end of the valid sequence with a five-frame transition and uses the last ten valid frames when no contact is predicted.
All \method{} variants use these fixed values.

All baselines are trained from scratch on the corresponding \dataset{} split (OpenHOI keeps its released affordance-tuned multimodal LLM),
using the optimization budgets of their original configurations where released.
\texthoi{} uses 30,000 generator, 10,000 contact-estimator, and 3,000 refiner iterations at batch size 64. DiffH2O uses 400,000 iterations, and
LatentHOI uses
15 epochs for its grasp VAE followed by 1,250 epochs for latent diffusion. The OpenHOI generator is trained for as many epochs as that of \texthoi{}. We change the
data interface and use the one-stage configuration of DiffH2O (\cref{sec:supp_crossmodal}), but do not shorten a
baseline merely to match our training time.

The diffusion Transformer has 8 layers with latent dimension $D=512$, feed-forward dimension 1,024, 4 attention heads, and dropout 0.1. For training stability, we apply a 10\% gaze-condition dropout, which replaces the whole gaze condition of a sample with the zero gaze token. The model is trained with AdamW~\citep{adamw} (learning rate $10^{-4}$, weight decay $10^{-2}$, batch size 64) under a 1,000-step cosine diffusion schedule~\citep{iddpm}, and sampling uses all 1,000 steps. Object points are normalized to $r_{\mathrm{norm}}=0.85$ within the unit-ball BPS domain ($K=1{,}024$).

The \method{} model is trained for 4,000 epochs on each fold, and the final checkpoint is used without validation-based selection. The model has 23.9M trainable parameters, and training on one fold takes 14 to 16 hours on one NVIDIA RTX 3090 (24\,GB).

The software environment is PyTorch 1.13.0 with CUDA 11.6 and PyTorch3D 0.7.2.

\begin{table*}[t]
\centering
\caption{\textbf{Metric protocol.} The metrics reported in the main paper and in the supplementary material. GSR is the primary outcome.}
\label{tab:metric_protocol}
\small
\setlength{\tabcolsep}{4pt}
\begin{tabular}{@{}>{\raggedright\arraybackslash}p{0.28\textwidth}>{\raggedright\arraybackslash}p{0.24\textwidth}>{\raggedright\arraybackslash}p{0.43\textwidth}@{}}
\toprule
Metric & Support & Meaning\\
\midrule
Grasp Success Rate (GSR) & Final five valid frames (contact), final valid frame (penetration, lift)
& \leavevmode{}Contact in each of the final five frames, final-frame penetration depth at most 10\,mm, and final lift of at least 5\,mm along the world-up direction\\
Part Contact Precision (PCP) / Part Accuracy (Part Acc.) & Final valid frame & Target-part precision and dominant-part match\\
\leavevmode{}Interpenetration Depth (ID) & \leavevmode{}Penetrating valid frames & \leavevmode{}Maximum depth (mm) per frame, averaged over penetrating frames\\
\leavevmode{}Contact & \leavevmode{}Final five valid frames & \leavevmode{}Proportion meeting the contact condition of GSR\\
\leavevmode{}Sim / SimD & \leavevmode{}Final valid frame & \leavevmode{}Object held in physics simulation (displacement below 2\,cm) / mean object displacement (cm)\\
\leavevmode{}Follow & \leavevmode{}Rotated input & \leavevmode{}Final-wrist azimuth about the object's up axis turns with the input rotation within 45$^\circ$\\
TSA / DCR / Held & Multi-object scene & Target-contact selectivity / distractor contact rate / target still held in the final five frames\\
\bottomrule
\end{tabular}
\end{table*}

\section{Metric definitions and aggregation}\label{sec:supp_metrics}

We use the annotated active hand(s), ignore padding and absent hands, and average
each metric over the samples of a fold and then over the three folds. \Cref{tab:metric_protocol} gives the
compact protocol. Differences between methods are tested with clip-level bootstrap 95\% confidence intervals (1,000--10,000 resamples), paired on clips wherever the outputs of two methods can be matched to the same clip. In the main comparison of \ifwithsupp\cref{tab:quantitative_results}\else Tab.~\suppref{tab:quantitative_results}{1} of the main paper\fi{}, the samples of every baseline are matched to the same clips as ours (same object, clip, and length), so all tests are paired. When objects instead of clips are resampled within each fold, the lead of \method{} over LatentHOI remains significant for GSR ($+19.7$, $[11.9,28.4]$) but not for Part Acc.\ ($+5.7$, $[-4.4,16.2]$) or PCP ($+7.6$, $[-1.5,17.2]$), whereas over \texthoi{} all three remain significant.

\begin{table}[t]
\centering
\caption{\textbf{Effect of the contact refinement} (three object folds, \%). Own pipeline: each baseline as released. MANO fit: \method{} without refinement. Nearest-pt.\ and predicted: the same 300-step refinement with displacements to the nearest object points or as predicted by \method{}. ID in mm. Metrics as in \cref{tab:metric_protocol}. Bold: best value.}
\label{tab:supp_shared_pp}
\footnotesize
\setlength{\tabcolsep}{0pt}
\begin{tabular*}{\columnwidth}{@{\extracolsep{\fill}}llcccc@{}}
\toprule
Method & Refinement & GSR\,$\uparrow$ & Part Acc.\,$\uparrow$ & PCP\,$\uparrow$ & ID\,$\downarrow$\\
\midrule
\texthoi{} & own pipeline & 35.3 & 36.8 & 36.6 & 12.0\\
 & + nearest-pt. & 40.3 & 37.8 & 38.0 & 10.2\\
DiffH2O & own pipeline & 17.9 & 31.7 & 31.0 & 8.9\\
 & + nearest-pt. & 22.7 & 33.1 & 32.5 & 7.4\\
LatentHOI & own pipeline & 46.6 & 43.6 & 42.5 & 8.4\\
 & + nearest-pt. & 55.6 & 44.8 & 44.1 & 7.2\\
OpenHOI & own pipeline & 25.8 & 31.5 & 31.0 & 11.9\\
 & + nearest-pt. & 28.3 & 33.6 & 33.6 & 10.8\\
\midrule
\method{} (Ours) & MANO fit & 40.6 & 37.6 & 38.6 & 9.6\\
 & + nearest-pt. & 47.3 & 40.1 & 41.2 & 8.1\\
 & + predicted & \textbf{66.3} & \textbf{49.3} & \textbf{50.1} & \textbf{6.7}\\
\bottomrule
\end{tabular*}
\end{table}

\begin{table*}[t]
\centering
\caption{\textbf{Per-fold results} (\%) for Tab.~\suppref{tab:quantitative_results}{1} of the main paper and \cref{tab:convey,tab:design,tab:supp_shared_pp}. Bold: best value of each block in each column.}
\label{tab:supp_perfold}
\small
\setlength{\tabcolsep}{0pt}
\begin{tabular*}{\textwidth}{@{\extracolsep{\fill}}lccccccccc@{}}
\toprule
& \multicolumn{3}{c}{Object fold 1} & \multicolumn{3}{c}{Object fold 2} & \multicolumn{3}{c}{Object fold 3}\\
\cmidrule(lr{1em}){2-4}\cmidrule(lr{1em}){5-7}\cmidrule(l{1em}){8-10}
Row & GSR\,$\uparrow$ & Part Acc.\,$\uparrow$ & PCP\,$\uparrow$ & GSR\,$\uparrow$ & Part Acc.\,$\uparrow$ & PCP\,$\uparrow$ & GSR\,$\uparrow$ & Part Acc.\,$\uparrow$ & PCP\,$\uparrow$\\
\midrule
\multicolumn{10}{@{}l}{\textit{Table~\suppref{tab:quantitative_results}{1} of the main paper (own pipelines)}}\\
\texthoi{} & 23.2 & 43.8 & 43.0 & 41.4 & 37.6 & 37.9 & 41.4 & 29.0 & 29.0\\
DiffH2O & 12.1 & 30.0 & 28.4 & 17.0 & 33.8 & 34.5 & 24.6 & 31.4 & 30.2\\
LatentHOI & 35.6 & 42.5 & 41.7 & 51.9 & 50.5 & 48.3 & 52.3 & 37.9 & 37.4\\
OpenHOI & 13.9 & 34.7 & 33.7 & 25.7 & 30.3 & 30.1 & 37.7 & 29.6 & 29.3\\
\method{} (Ours) & \textbf{53.6} & \textbf{56.0} & \textbf{56.9} & \textbf{68.8} & \textbf{51.7} & \textbf{51.1} & \textbf{76.6} & \textbf{40.2} & \textbf{42.1}\\
\midrule
\multicolumn{10}{@{}l}{\textit{\Cref{tab:supp_shared_pp} (MANO fit and nearest-point refinement)}}\\
\texthoi{} & 27.4 & 45.2 & 44.4 & 44.5 & 38.1 & 38.6 & 49.1 & 30.2 & 31.1\\
DiffH2O & 16.3 & 30.7 & 29.8 & 21.7 & 35.8 & 35.8 & 30.2 & 32.8 & 32.0\\
LatentHOI & \textbf{46.5} & 45.1 & 43.6 & \textbf{57.8} & \textbf{50.2} & \textbf{48.5} & \textbf{62.4} & \textbf{39.2} & \textbf{40.1}\\
OpenHOI & 14.6 & 36.3 & 35.5 & 27.4 & 33.4 & 33.6 & 42.8 & 31.0 & 31.6\\
\method{}, MANO fit & 31.6 & 46.5 & 47.0 & 41.9 & 36.2 & 37.1 & 48.4 & 30.0 & 31.8\\
\method{}, MANO fit + nearest-pt. & 38.0 & \textbf{48.5} & \textbf{49.7} & 47.1 & 38.8 & 39.3 & 56.7 & 33.0 & 34.7\\
\midrule
\multicolumn{10}{@{}l}{\textit{\Cref{tab:convey} (same backbone)}}\\
No condition & 47.3 & 48.4 & 47.8 & 59.1 & 45.4 & 45.7 & 70.7 & 37.2 & 37.3\\
Part label & 49.7 & 55.5 & 53.6 & 65.7 & 51.0 & 50.3 & 67.0 & 39.9 & 41.2\\
Text prompt & 55.6 & 55.3 & 52.6 & 68.4 & \textbf{54.3} & 53.2 & 70.5 & \textbf{47.0} & \textbf{46.1}\\
Gaze point & 45.2 & 52.8 & 52.2 & 61.7 & 46.4 & 46.8 & 72.7 & 40.2 & 41.4\\
Head-only & 47.8 & 53.1 & 54.1 & 64.2 & 49.7 & 48.7 & \textbf{77.3} & 41.1 & 43.3\\
Raw ray & \textbf{57.4} & 53.8 & 53.5 & 67.5 & 53.4 & \textbf{53.3} & 75.7 & 44.1 & 45.5\\
GHO-style & 53.5 & 58.8 & 57.7 & 70.3 & 49.6 & 50.9 & 75.9 & 43.5 & 45.5\\
\geogaze{} & 53.6 & 56.0 & 56.9 & 68.8 & 51.7 & 51.1 & 76.6 & 40.2 & 42.1\\
\geogaze{} $+$ text prompt & 51.2 & \textbf{59.7} & \textbf{59.8} & \textbf{72.1} & 52.8 & 51.8 & 74.6 & 42.3 & 45.3\\
\midrule
\multicolumn{10}{@{}l}{\textit{\Cref{tab:design} (gaze encoding)}}\\
Ray closeness only & 42.2 & 51.5 & 52.6 & 59.1 & 47.2 & 47.4 & 69.6 & \textbf{41.5} & 42.2\\
Alignment only & \textbf{58.0} & 55.5 & 55.0 & \textbf{71.0} & \textbf{52.3} & \textbf{52.2} & 76.0 & 39.7 & \textbf{43.1}\\
Both (Ours) & 53.6 & \textbf{56.0} & \textbf{56.9} & 68.8 & 51.7 & 51.1 & \textbf{76.6} & 40.2 & 42.1\\
\bottomrule
\end{tabular*}
\end{table*}

\paragraph{Penetration and contact-based success.}
A valid frame $i$ of a sample is \emph{in contact} when at least one vertex of the annotated active hand(s) lies within 5\,mm of its nearest object-mesh vertex. This is a nearest-vertex
approximation rather than an exact point-to-triangle distance. The penetration depth $\Delta_{i}$ is the depth of the deepest active-hand vertex inside the object, with $\Delta_{i}=0$ when no vertex is inside. ID averages $\Delta_{i}$ over the penetrating frames. Depths no greater
than 2\,mm are treated as zero, and samples without a penetrating frame are omitted. A prediction that never touches the object therefore has no penetration to count, so ID is read together with the contact-based metrics. The same ID is reported on GazeHOI (Tab.~\suppref{tab:gazehoi}{2} of the main paper and \cref{tab:supp_gazehoi_pp}).

Let $\mathcal W$ be the final five valid frames, $i^\star$ the final valid
frame, and $L_{i}$ the displacement of the object bottom from its
initial-frame height along the HOT3D world-up direction $\mathbf u$ (the lowest object vertex projected on $\mathbf u$). A sample passes Contact when every frame of $\mathcal W$ is in contact, and
\begin{equation}
\mathrm{GSR}
=\mathrm{Contact}\,
\mathbf 1[\Delta_{i^\star}\leq 10\,\mathrm{mm}]\,
\mathbf 1[L_{i^\star}\geq 5\,\mathrm{mm}].
\end{equation}
A sequence with fewer than five valid frames
fails Contact and therefore GSR.

\paragraph{Simulation stability (Sim, SimD).}\label{sec:supp_sim}
We follow the simulation-displacement protocol of ObMan~\citep{obman}, reused by GraspTTA~\citep{contactgrasp}, and port its reference PyBullet~\citep{pybullet} settings unchanged. Each sample is simulated once at its final valid frame, the frame at which GSR is evaluated, with the hand and object in the evaluator's world frame. The final MANO hand mesh is a static, massless concave triangle mesh. The object is a dynamic body built from a V-HACD~\citep{vhacd} convex decomposition of its mesh, with a mass of 1\,kg and its center of mass at the mean of the mesh vertices. Gravity of 9.8\,m/s$^2$ acts against the HOT3D world-up direction. The simulation runs for 35 steps of 1/240\,s (0.146\,s), and no table or other support is present. SimD is the distance between the initial and final object center of mass, averaged over all samples. Sim is the proportion of samples displaced by less than 2\,cm. An object that touches no hand falls freely, about 10.7\,cm in this duration, and is kept in both averages. ObMan notes that balanced penetration forces can also hold an object, so Sim and SimD are read together with ID. Among the samples in contact whose final-frame penetration depth is at most 10\,mm, \method{} holds the object in 80.8\% of the cases, against at most 70.4\% for the baselines, whereas above 10\,mm LatentHOI is slightly more stable (67.8\% vs.\ 63.1\%). Run on the recorded motions, the same protocol gives Sim 73.9\% and SimD 2.04\,cm over the three object folds, which serves as a reference for the generated grasps.

\paragraph{Part metrics.}
PCP maps each final contacting hand vertex to its nearest object point,
deduplicates these points, and measures the fraction of them on the annotated target part. Part Accuracy is one when the target
part has the largest such contact count and zero otherwise. A sample with no
final contact receives zero for both part metrics. \Cref{tab:supp_partmacro} also reports both metrics without the one-part objects and as macro averages.

\paragraph{Multi-object target-contact selectivity.}
For multi-object evaluation, an object is counted as contacted when at least
two active-hand keypoints lie within 2\,cm in any valid frame. TSA measures target-contact selectivity rather than stable grasping. The 2\,cm threshold
compensates for sparse
keypoint and mesh discretization and is applied identically to all methods.

\section{Additional quantitative results}

\subsection{Nearest-point refinement and per-fold results}\leavevmode%
Table~1 of the main paper evaluates each baseline with its released pipeline, and \cref{tab:supp_shared_pp} adds the nearest-point refinement to every baseline and separates its effect. With nearest-point displacements, this refinement raises GSR by 2.5 to 9.0 points for every baseline and by 6.7 points for the MANO fit of \method{}. It raises Part Acc.\ by 1.0 to 2.5 points for all five. \method{}'s own refinement adds 25.7 GSR and 11.7 Part Acc.\ points to the same MANO fit. The difference comes from what the network predicts: \method{} outputs sampled hand vertices with their displacement vectors to the object instead of MANO parameters, and its refinement uses these predicted displacements as contact targets, whereas the nearest-point version sees only the fitted mesh. With nearest-point displacements on both, LatentHOI exceeds \method{} by 8.3 GSR and 4.7 Part Acc.\ points (95\% CIs $[5.0,11.5]$ and $[1.6,7.9]$), so the results of \method{} rely on its predicted displacements. The refinement also lowers ID in every row. \Cref{tab:supp_perfold} lists the per-fold values behind Tab.~\suppref{tab:quantitative_results}{1} of the main paper and \cref{tab:supp_shared_pp,tab:convey,tab:design}.

\begin{table}[t]
\centering
\caption{\textbf{Zero-shot transfer to GazeHOI with the nearest-point refinement} (939 clips, one sample per clip, \%, ID in mm, mean of the three object-fold models). Refinement rows as in \cref{tab:supp_shared_pp}. Metrics as in \cref{tab:metric_protocol}. Bold: best value.}
\label{tab:supp_gazehoi_pp}
\small
\setlength{\tabcolsep}{0pt}
\begin{tabular*}{\columnwidth}{@{\extracolsep{\fill}}llcccc@{}}
\toprule
Method & Refinement & GSR\,$\uparrow$ & Sim\,$\uparrow$ & Contact\,$\uparrow$ & ID\,$\downarrow$\\
\midrule
\texthoi{} & own pipeline & 18.4 & 27.1 & 46.8 & 11.2\\
 & + nearest-pt. & 24.9 & 31.9 & 51.9 & \textbf{9.5}\\
DiffH2O & own pipeline & 11.1 & 22.7 & 35.3 & 12.8\\
 & + nearest-pt. & 15.1 & 25.7 & 38.7 & 10.8\\
LatentHOI & own pipeline & 33.8 & 48.6 & 72.6 & 11.6\\
 & + nearest-pt. & 40.0 & 52.3 & 76.3 & 10.3\\
OpenHOI & own pipeline & 16.8 & 34.0 & 51.8 & 13.3\\
 & + nearest-pt. & 18.5 & 38.0 & 53.3 & 12.0\\
\midrule
\method{} (Ours) & + predicted & \textbf{45.3} & \textbf{67.2} & \textbf{92.2} & 10.3\\
\bottomrule
\end{tabular*}
\end{table}

\Cref{tab:supp_gazehoi_pp} repeats the zero-shot GazeHOI comparison of Tab.~\suppref{tab:gazehoi}{2} of the main paper with the nearest-point refinement applied to every baseline. The refinement raises GSR by 1.7 to 6.5 points. LatentHOI then reaches 40.0\%, still significantly below \method{} (45.3\%, difference 5.3 points, 95\% CI $[3.0,7.7]$), and \method{} keeps a significantly higher final-contact rate (92.2\% vs.\ 76.3\%) and the most stable grasps in simulation (Sim 67.2\% vs.\ at most 52.3\%, paired difference to LatentHOI $+14.8$ points, $[12.5,17.3]$). The refinement lowers the ID of every baseline by 1.3 to 2.0\,mm, to 9.5\,mm for \texthoi{} and 10.3\,mm for LatentHOI (\method{}: 10.3\,mm).

\begin{table}[t]
\centering
\caption{\textbf{Same backbone, different conditioning} (three object folds, \%, refinement as in Tab.~\suppref{tab:quantitative_results}{1} of the main paper). Rows differ only in the condition (\cref{sec:supp_rotation}). Follow: proportion of rotated-gaze inputs, on the 848 single-hand clips, whose final wrist turns with the rotation within 45$^\circ$ (Sec.~\suppref{sec:exp_convey}{4.4} of the main paper). --: no gaze to rotate. Bold: best value.}
\label{tab:convey}
\small
\setlength{\tabcolsep}{0pt}
\begin{tabular*}{\columnwidth}{@{\extracolsep{\fill}}lcccc@{}}
\toprule
Condition & GSR\,$\uparrow$ & Part Acc.\,$\uparrow$ & PCP\,$\uparrow$ & Follow\,$\uparrow$\\
\midrule
\multicolumn{5}{@{}l}{\footnotesize\textit{Without gaze}}\\
\quad No condition & 59.0 & 43.7 & 43.6 & --\\
\quad Part label & 60.8 & 48.8 & 48.4 & --\\
\quad Text prompt & 64.8 & \textbf{52.2} & 50.6 & --\\
\multicolumn{5}{@{}l}{\footnotesize\textit{Derived from the recorded gaze}}\\
\quad Gaze point & 59.9 & 46.5 & 46.8 & 28.6\\
\quad Head-only & 63.1 & 48.0 & 48.7 & 54.6\\
\quad Raw ray & \textbf{66.9} & 50.4 & 50.8 & 50.8\\
\quad GHO-style & 66.6 & 50.6 & 51.4 & 53.7\\
\quad \geogaze{} (Ours) & 66.3 & 49.3 & 50.1 & \textbf{59.1}\\
\multicolumn{5}{@{}l}{\footnotesize\textit{Gaze and text}}\\
\quad \geogaze{} $+$ text prompt & 66.0 & 51.6 & \textbf{52.3} & 57.4\\
\bottomrule
\end{tabular*}
\end{table}

\begin{table}[t]
\centering
\caption{\textbf{Gaze encoding analysis} (three object folds, refinement as in Tab.~\suppref{tab:quantitative_results}{1} of the main paper, \%). Rows differ only in the gaze fields given to the model. Follow: as in \cref{tab:convey}. Both is the model of Tab.~\suppref{tab:quantitative_results}{1} of the main paper. Follow uses four samples per clip. Bold: best value.}
\label{tab:design}

\small
\setlength{\tabcolsep}{0pt}
\begin{tabular*}{\columnwidth}{@{\extracolsep{\fill}}lccc@{}}
\toprule
Variant & GSR\,$\uparrow$ & Part Acc.\,$\uparrow$ & Follow\,$\uparrow$\\
\midrule
Ray closeness only & 57.0 & 46.7 & 37.9\\
Alignment only & \textbf{68.3} & 49.2 & 53.6\\
Both (Ours) & 66.3 & \textbf{49.3} & \textbf{59.1}\\
\bottomrule
\end{tabular*}
\end{table}

\subsection{Gaze conditions, rotation, and clock words}\label{sec:supp_rotation}

\paragraph{Same-backbone conditions.}The rows of \cref{tab:convey} differ only in the condition. Text prompt is the sentence the text baselines receive (\eg, ``Grab cap of bottle\_bbq with right hand.''), encoded by the frozen CLIP~\citep{clip} text encoder, and Part label gives the ground-truth part as input. Raw ray gives the gaze origin and direction unchanged. Gaze point keeps, per frame, only the object point nearest the gaze ray, without eye position or direction. Head-only is trained and tested with only the ray from the eye to the object centroid, without the gaze direction. GHO-style, a simplified input after GHO-Diffusion~\citep{gazehoi}, gives the eye ray with the object point nearest to it and their distance, without the inverse-distance weighting and self-attention aggregation of GHO-Diffusion. With the rotated gaze of Sec.~\suppref{sec:exp_convey}{4.4} of the main paper, Head-only, the raw ray, and the GHO-style feature all follow the rotation less often than \geogaze{} (54.6\%, 50.8\%, and 53.7\% vs.\ 59.1\%, all significant). Head-only already improves over no condition (GSR $+4.1$ $[1.3,6.8]$, Part Acc.\ $+4.3$ $[2.1,6.5]$, PCP $+5.1$ $[3.3,6.9]$, all significant). \geogaze{} leads Head-only in GSR ($+3.2$, $[0.2,6.3]$, significant) but not in Part Acc.\ ($+1.3$, $[-1.3,4.0]$) or PCP ($+1.3$, $[-0.8,3.4]$), both not significant. The gaze direction matters when the gaze leaves the object center: on the 11\% of clips (99) whose gaze deviates from the eye-to-centroid ray by at least 10$^\circ$, \geogaze{} reaches a 19-point higher GSR than Head-only (95\% CI $[8.2,29.3]$, 2,000 bootstrap resamples), with no significant change in Part Acc.

On caps, rims, tops, and lids (219 clips), the text prompt reaches a Part Acc.\ of 34.7\%, against 19.6\% with gaze, because people also look at the top of an upright object in 84\% of the body grasps. Before first contact, the recorded gaze falls on the target part in 38.3\% of the frames, against 19.8\% for random fixations on the object from the same eye position. Adding the text prompt to \geogaze{} on the same backbone (last row of \cref{tab:convey}) matches the prompt alone in Part Acc.\ ($-0.6$, n.s.) and GSR ($+1.2$, n.s.) and makes the approach follow a rotated gaze (57.4\% of the rotated inputs), which the prompt alone cannot.

\begin{figure*}[t]
\centering
\includegraphics[width=\textwidth]{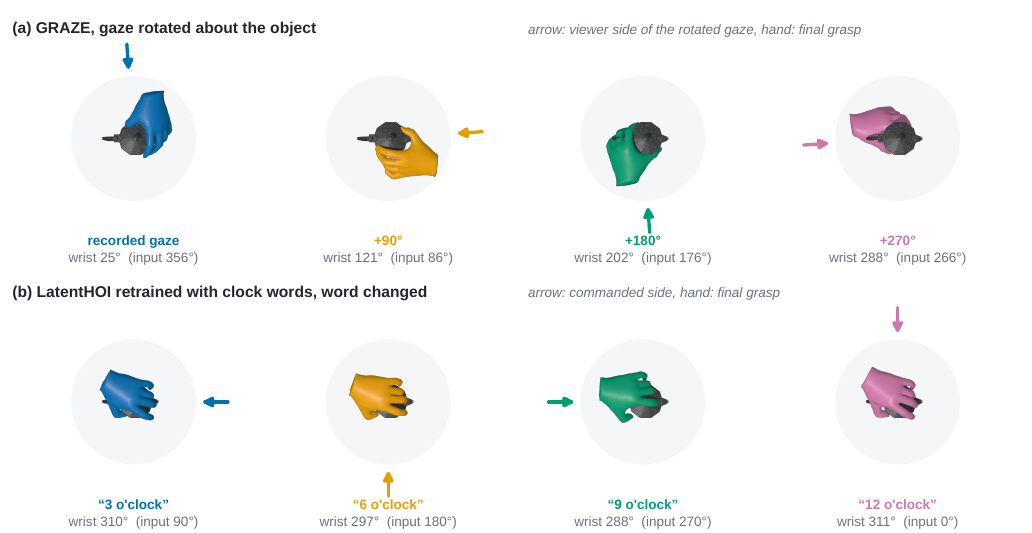}
\caption{\textbf{Rotating the input of one clip} (unseen coffee pot, top view). (a)~\method{} with the recorded gaze rotated about the object. (b)~LatentHOI retrained with clock words, with the word changed. Arrows: (a) viewer side of the rotated gaze, (b) side named by the clock word. Hands: final frame of the grasp generated for each input, drawn in the color of its arrow, with its wrist azimuth below. Colors encode the rotation step (blue: unrotated input), and +90$^\circ$ is clockwise from above.}
\label{fig:rot_example}
\end{figure*}

\paragraph{Gaze encoding.}\Cref{tab:design} separates the two \geogaze{} fields. Ray closeness alone lowers GSR by 9.4 points (95\% CI $[6.1,12.5]$) and Follow by 21.2 points ($[18.6,23.9]$): it marks where the ray meets the object, but not from which side it comes. Alignment alone reaches a similar GSR ($+2.0$, n.s.) and Part Acc.\ (n.s.), but follows the rotation 5.5 points less often ($[3.2,7.9]$). Combining the two fields gives the highest Follow at the same Part Acc., which is why \method{} uses both.

\begin{table}[!t]
\centering
\caption{\textbf{Gaze corruption at test time} (three object folds). Change from the uncorrupted gaze in percentage points, with clip-level bootstrap 95\% CIs. All intervals include zero.}
\label{tab:supp_gaze_noise}
\small
\setlength{\tabcolsep}{0pt}
\begin{tabular*}{\columnwidth}{@{\extracolsep{\fill}}lcc@{}}
\toprule
Corruption & $\Delta$GSR & $\Delta$Part Acc.\\
\midrule
Angular noise, 5$^\circ$ & $+0.0$ $[-2.0, 2.0]$ & $-1.1$ $[-2.8, 0.6]$\\
Angular noise, 10$^\circ$ & $+0.1$ $[-1.9, 2.1]$ & $-0.6$ $[-2.3, 1.1]$\\
Frame drop, 20\% & $+0.1$ $[-0.6, 0.8]$ & $-0.1$ $[-0.6, 0.5]$\\
Frame drop, 50\% & $-0.5$ $[-1.6, 0.7]$ & $-0.3$ $[-1.2, 0.6]$\\
\bottomrule
\end{tabular*}
\end{table}

\begin{table*}[t]
\centering
\caption{\textbf{Part metrics} (\%, three object folds) for the rows of Tab.~\suppref{tab:quantitative_results}{1} of the main paper and \cref{tab:supp_shared_pp}. \textit{All}: all objects, averaged over the three folds as in Tab.~\suppref{tab:quantitative_results}{1} of the main paper. \textit{Micro}: the same without the five one-part objects. \textit{Obj.}\ and \textit{Pair}: macro averages over objects and over object--part pairs, without the one-part objects. Bold: best.}
\label{tab:supp_partmacro}
\small
\setlength{\tabcolsep}{0pt}
\begin{tabular*}{\textwidth}{@{\extracolsep{\fill}}lcccccccc@{}}
\toprule
& \multicolumn{4}{c}{Part Acc.\,$\uparrow$} & \multicolumn{4}{c@{}}{PCP\,$\uparrow$}\\
\cmidrule(lr{0.6em}){2-5}\cmidrule(l{0.6em}){6-9}
Method & All & Micro & Obj. & Pair & All & Micro & Obj. & Pair\\
\midrule
\multicolumn{9}{@{}l}{\footnotesize\textit{Own pipeline}}\\
\hspace{0.7em}\texthoi{} & 36.8 & 33.0 & 36.6 & 38.2 & 36.6 & 32.8 & 36.1 & 37.7\\
\hspace{0.7em}DiffH2O & 31.7 & 28.8 & 30.5 & 29.9 & 31.0 & 28.0 & 29.6 & 28.9\\
\hspace{0.7em}LatentHOI & 43.6 & 39.2 & 41.1 & 42.0 & 42.5 & 37.9 & 40.1 & 40.7\\
\hspace{0.7em}OpenHOI & 31.5 & 29.5 & 31.8 & 34.5 & 31.0 & 28.9 & 31.1 & 34.1\\
\multicolumn{9}{@{}l}{\footnotesize\textit{+ nearest-pt.}}\\
\hspace{0.7em}\texthoi{} & 37.8 & 33.8 & 37.4 & 39.6 & 38.0 & 33.9 & 37.2 & 39.4\\
\hspace{0.7em}DiffH2O & 33.1 & 29.8 & 31.9 & 31.3 & 32.5 & 29.2 & 31.0 & 30.2\\
\hspace{0.7em}LatentHOI & 44.8 & 40.3 & 42.3 & 43.3 & 44.1 & 39.3 & 41.6 & 42.0\\
\hspace{0.7em}OpenHOI & 33.6 & 30.9 & 33.8 & 35.9 & 33.6 & 30.9 & 33.5 & 36.3\\
\midrule
\method{} (Ours) & \textbf{49.3} & \textbf{43.2} & \textbf{46.9} & \textbf{44.7} & \textbf{50.1} & \textbf{44.0} & \textbf{47.2} & \textbf{45.4}\\
\bottomrule
\end{tabular*}
\end{table*}

\paragraph{Gaze noise.}\Cref{tab:supp_gaze_noise} corrupts the recorded gaze at test time for the model of Tab.~\suppref{tab:quantitative_results}{1} of the main paper, without retraining: random angular noise on every gaze direction (RMS error of 5$^\circ$ or 10$^\circ$), or each gaze frame dropped with probability 0.2 or 0.5, which zeroes its gaze token. No corruption changes GSR or Part Acc.\ significantly.

\paragraph{Clock words and rotated inputs.}Clock words are defined in the frame of each object, so that they need no world pose. The up axis is the signed mesh axis ($\pm x$, $\pm y$, or $\pm z$) that points closest to the world-up direction in most clips of the object. Twelve o'clock is the positive direction of the next mesh axis in the cyclic order $x \to y \to z \to x$, and 3~o'clock lies 90$^\circ$ clockwise from it when seen from above. The recorded direction of a clip is the azimuth of the active-hand center at the first contact frame around this up axis, assigned to the nearest of the four clock positions. LatentHOI and our backbone with the text prompt are retrained on every object fold with ``, approaching from N o'clock'' appended to each training prompt, where N is the recorded direction. Clips in which both hands act are excluded. Follow is computed on the remaining 848 single-hand clips of the three object folds for every method, with four samples per clip. At test time, every clip receives the recorded word and the three words rotated by 90$^\circ$, 180$^\circ$, and 270$^\circ$. For \method{}, the recorded gaze rays are instead rotated by the same angles about the up axis of the object through its first-frame centroid. An output follows the rotation when its final-wrist azimuth, relative to the output for the unrotated input with the same sampling noise, has turned by the rotation angle to within $\pm$45$^\circ$. Fig.~\suppref{fig:control}{5}b of the main paper further splits the remaining outputs: one that stayed within 45$^\circ$ of the output for the unrotated input did not turn, and every other one turned elsewhere. Classifier-free guidance at inference (scale 2.5, as in LatentHOI) does not make our backbone follow the clock word: it follows the rotation in 2.2\% of the rotated inputs with and without guidance, although guidance changes the outputs (median final-wrist change 14--19$^\circ$ across the three folds).

\paragraph{Rotated input on one clip.}\Cref{fig:rot_example} shows one clip of object fold 1 from above, chosen among the clips that \method{} follows on all three rotations. Rotating the recorded gaze by 90$^\circ$, 180$^\circ$, and 270$^\circ$ turns the final wrist of \method{} by 96$^\circ$, 177$^\circ$, and 263$^\circ$, each within 7$^\circ$ of the rotation. Changing the clock word of LatentHOI from 3 to 6, 9, and 12 o'clock moves its final wrist by $-13^\circ$, $-22^\circ$, and $+1^\circ$, so the wrist stays on the same side of the object whichever side the word names.

\paragraph{Left and right words.}For Fig.~\suppref{fig:task_overview}{1}b of the main paper, the LatentHOI model of object fold 2 receives ``Grab right head of dumbbell'' and ``Grab left head of dumbbell'' on the 15 dumbbell clips, with ten samples per clip and prompt. All 183 samples that touch a head touch the same end of the mesh (60 for ``left'', 123 for ``right''), and the other end is never touched. The word therefore does not select the head. These prompts are a probe: the benchmark target of all 15 dumbbell clips is the center bar.

\begin{table}[!t]
\centering
\caption{\textbf{Subject-disjoint folds} (unseen subjects, every test object is seen in training, \%, ID in mm). Mean $\pm$ sample SD over the three folds, whose test sets cover the 912 clips. Sim is not computed. Bold: best.}
\label{tab:supp_main_table1_sd}
\footnotesize
\setlength{\tabcolsep}{0pt}
\begin{tabular*}{\columnwidth}{@{\extracolsep{\fill}}lcccc@{}}
\toprule
Method & GSR\,$\uparrow$ & Part Acc.\,$\uparrow$ & PCP\,$\uparrow$ & ID\,$\downarrow$\\
\midrule
\texthoi{} & 56.8$\pm$3.9 & \textbf{71.7}$\pm$2.7 & \textbf{69.4}$\pm$2.1 & 8.0$\pm$0.8\\
DiffH2O & 26.0$\pm$5.3 & 60.9$\pm$3.9 & 58.5$\pm$3.3 & \textbf{5.9}$\pm$0.5\\
LatentHOI & 43.1$\pm$14.2 & 69.7$\pm$6.6 & 67.4$\pm$5.5 & 6.2$\pm$0.1\\
OpenHOI & 21.9$\pm$1.9 & 41.6$\pm$1.1 & 40.9$\pm$0.7 & 8.5$\pm$1.1\\
\midrule
\method{} (Ours) & \textbf{69.7}$\pm$3.1 & 59.2$\pm$3.3 & 59.7$\pm$3.0 & 6.2$\pm$0.2\\
\bottomrule
\end{tabular*}
\end{table}

\paragraph{More detailed prompts for the text baseline.}The prompts of the text baselines are short, which could put them at a disadvantage against a per-frame gaze input. DiffH2O~\citep{diffh2o} shows that more detailed text descriptions give finer control over the generated interaction. Its descriptions detail the sequence of actions. Our task, however, is a single grasp, so we instead detail how the hand grips the object. We retrain LatentHOI with such a prompt: each prompt gets an added sentence that names the fingers holding the object, derived from the recorded contact of the same clip (\eg, ``The thumb and index finger pinch it.''). The added detail does not help: on the three object folds (own pipeline, as in Tab.~\suppref{tab:quantitative_results}{1} of the main paper), GSR is unchanged (46.6\% vs.\ 46.6\%, 95\% CI of the difference $[-2.1,2.1]$), and Part Acc.\ and PCP fall (41.6\% vs.\ 43.6\% and 40.1\% vs.\ 42.5\%).

\begin{figure}[t]
\centering
\includegraphics[width=\columnwidth]{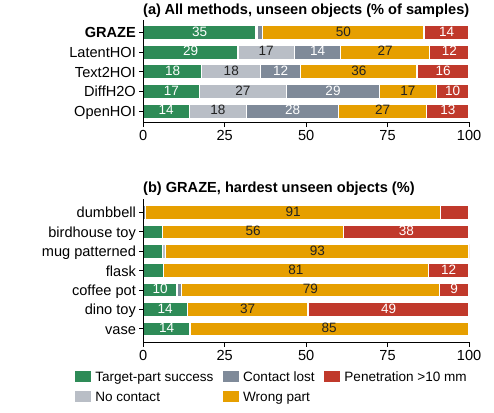}
\caption{\textbf{Failure taxonomy on unseen objects} (three object folds,
all samples). Each sample is assigned to one outcome. (a) All methods.
(b) The seven unseen objects on which \method{} has the lowest target-part
success.}
\label{fig:failure_modes}
\end{figure}

\begin{figure*}[t]
\centering
\includegraphics[width=\textwidth]{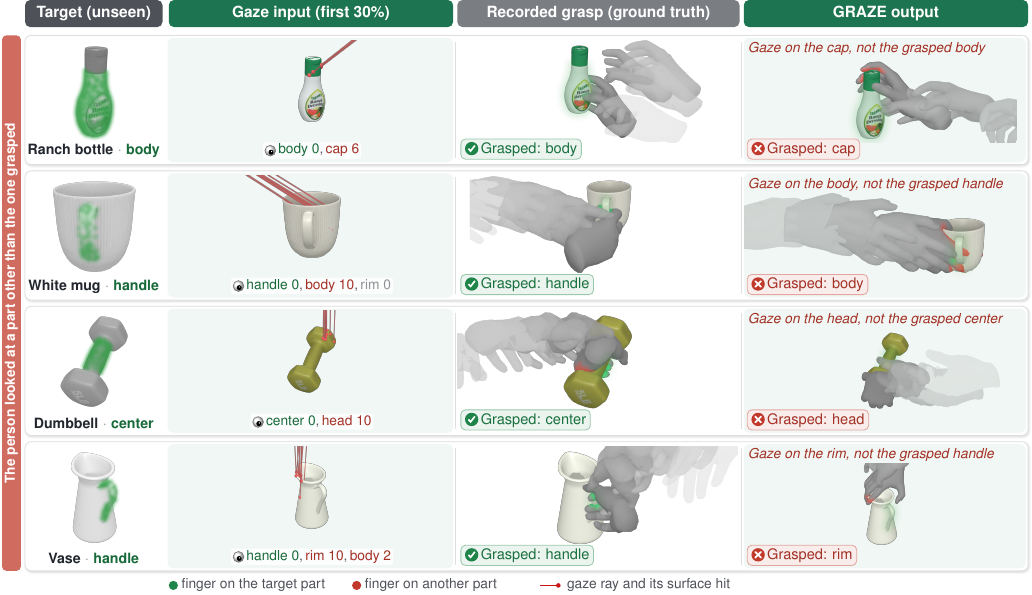}
\caption{\textbf{Wrong-part grasps of \method{} on unseen objects, in clips where the person looked at a part other than the one grasped.} Columns: the target part (green), the gaze that \method{} receives (first 30\% of the clip) with its hits per part, the recorded grasp, and the output of \method{}. Fingers near the object are green on the target part and red elsewhere. Labels: the part with most of the final contact (green: recorded grasp, red: \method{}). Gaze rays end with a dot where they meet the object. Numbers count these hits per part. Scales differ between cells.}
\label{fig:gaze_failures}
\end{figure*}

\subsection{Subject-disjoint folds and part metrics}

\leavevmode\Cref{tab:supp_main_table1_sd} reports the subject-disjoint folds, where every test object is seen in training: \method{} attains the highest GSR, and \texthoi{} the highest Part Acc.\ and PCP.

\paragraph{Part metrics without one-part objects.}Five of the 31 objects have a single annotated part. On their 103 clips, Part Acc.\ and PCP are one whenever the final grasp touches the object, so these clips measure contact and not the choice of part. \Cref{tab:supp_partmacro} therefore recomputes both metrics for the rows of Tab.~\suppref{tab:quantitative_results}{1} of the main paper without these objects (809 clips, 26 objects, and 54 object--part pairs over the three folds) in three ways: averaged over clips (Micro), over objects (Obj.), and over object--part pairs (Pair). Removing the one-part objects lowers the clip average of every method by 2.0 to 6.1 points, and \method{} keeps the highest value in every column. Under an object-level bootstrap, \method{} significantly exceeds \texthoi{} with the nearest-point refinement in five of the six Micro, Obj., and Pair columns, whereas its margin over LatentHOI with the nearest-point refinement (1.4 to 5.6 points) is not significant.

\section{Failure modes and limitations}\label{sec:supp_failure}

\paragraph{Failure taxonomy on unseen objects.}\leavevmode%
\Cref{fig:failure_modes} assigns every generated sample of the three object folds
to exactly one outcome, checked in this order: no contact (the hand never touches the object),
contact lost (the hand touches the object but not in all of the final five frames), final contact on a non-target part,
final-frame penetration deeper than 10\,mm, and target-part success (contact with the target
part throughout the final five frames and final-frame penetration depth $\leq$10\,mm). \method{} reaches the object far more reliably than
the text-conditioned baselines, but, once the target part is required, its dominant failure is a grasp on the wrong
part. On the dumbbell, the flask, the patterned mug, and the vase almost every failure is such a grasp, and penetration concentrates on the birdhouse toy and the dino toy. In 68\% of the wrong-part grasps of \method{}, the recorded gaze before first contact falls mostly on a part other than the target, against 32\% for its target-part successes.

\paragraph{Why \method{} grasps the wrong part.}\leavevmode%
\Cref{fig:gaze_failures} shows wrong-part grasps of \method{} with the gaze it received and the recorded grasp. In each case the person looked at one part while grasping another (the cap of the ranch bottle while taking its body, the body of the white mug while taking its handle, the head of the dumbbell while taking its center bar, and the rim of the vase while taking its handle), and \method{} followed the gaze.

\paragraph{Scope.}
\method{} is trained on single-object grasp-and-lift clips from nine HOT3D subjects (912 clips, 31 objects) and is tested outside HOT3D only zero-shot on GazeHOI. Motions beyond grasping are not covered. Wrong-part grasps remain the main failure mode (\cref{fig:failure_modes}), and the generator, trained from motion data without a physics simulator, can still form contact through penetration, which ID and Sim measure. Clip boundaries and part labels are manual annotations, and the study is offline, without a user evaluation.

\paragraph{Gaze quality and accessibility.}
The evaluation uses recorded gaze from HOT3D and GazeHOI and does not cover gaze-tracking
latency or a user in a closed loop with the generator.
Gaze should also be treated as an optional interface rather than a universal
replacement for language or other accessibility-oriented controls.

\ifdefined\FloatBarrier\FloatBarrier\fi